\documentclass[10pt,journal,compsoc]{IEEEtran}

\usepackage{graphicx,subfigure}
\usepackage{epstopdf}
\usepackage{bm}
\usepackage{amsmath,amssymb} 
\usepackage{color}
\usepackage{tabularx}
\usepackage{multirow}
\usepackage{bbm}
\usepackage{mathtools}
\usepackage{array}
\usepackage{algorithm}
\usepackage[noend]{algpseudocode}
\usepackage{pifont}
\newcommand{\cmark}{\ding{51}}%
\newcommand{\xmark}{\ding{55}}%

\newcommand{\changed}[1]{\textcolor{black}{#1}}

\ifCLASSOPTIONcompsoc
  \usepackage[nospace,nocompress]{cite}
\else
  \usepackage{cite}
\fi
\ifCLASSINFOpdf
\else
\fi
\begin{document}
%
\title{Unsupervised Tracklet Person Re-Identification}

\author{Minxian~Li,~\IEEEmembership{}
           Xiatian~Zhu,~\IEEEmembership{}
           and~Shaogang~Gong~\IEEEmembership{}
\IEEEcompsocitemizethanks{\IEEEcompsocthanksitem 
Minxian Li is with the School of Electronic Engineering and Computer Science, Queen Mary University of London, London E1 4NS, UK. E-mail: m.li@qmul.ac.uk.
\IEEEcompsocthanksitem
Xiatian Zhu is with Vision Semantics Limited, London E1 4NS, UK. E-mail: eddy@visionsemantics.com.
\IEEEcompsocthanksitem
Shaogang Gong is with the School of Electronic Engineering and Computer Science, Queen Mary University of London, London E1 4NS, UK.
E-mail: s.gong@qmul.ac.uk.} 
}

%
%

\markboth{
	Accepted to appear in IEEE Transactions on Pattern Analysis and Machine Intelligence
}%
{Li \MakeLowercase{\textit{et al.}}: Unsupervised Person Re-Identification}
%



\IEEEtitleabstractindextext{%
\begin{abstract}
Most existing person re-identification (re-id) methods rely on 
{\em supervised} model learning on per-camera-pair {\em manually} labelled pairwise training data.
This leads to poor scalability in a practical re-id deployment,
due to the lack of exhaustive identity labelling of positive and negative image pairs
for every camera-pair. 
In this work, we present an unsupervised re-id deep learning approach.
It is capable of incrementally discovering and exploiting the underlying re-id discriminative
information from {\em automatically} generated person tracklet data 
{\em end-to-end}.
We formulate an {\em Unsupervised Tracklet Association Learning} 
(UTAL) framework.
This is by 
jointly learning within-camera tracklet discrimination
and cross-camera tracklet association in order to maximise the discovery of 
tracklet identity matching 
both within and across camera views.
Extensive experiments 
demonstrate the superiority of the proposed model
over the state-of-the-art unsupervised learning and domain adaptation person re-id
methods on eight benchmarking datasets.
\end{abstract}

}

\maketitle

\IEEEdisplaynontitleabstractindextext

%
\IEEEpeerreviewmaketitle

\IEEEraisesectionheading{\section{Introduction}\label{sec:introduction}}

\IEEEPARstart{P}{erson} re-identification (re-id) aims to match the underlying identity classes 
of person bounding box images detected from
non-overlapping camera views \cite{gong2014person}.
In recent years, 
extensive research has been carried out on re-id
\cite{li2017person,li2018harmonious,wei2018person,song2018mask,chang2018multi,shen2018deep}. 
Most existing person re-id methods, in particular neural network deep learning models,
adopt the {\em supervised learning} approach. 
Supervised deep models assume the availability of a large number of {\em manually}
labelled {\em cross-view identity matching image pairs} for each 
camera pair.
This enables deriving a feature representation and/or a
distance metric function optimised for each camera-pair.
Such an assumption is inherently limited for generalising a person re-id model
to many different camera networks.
This is because, exhaustive manual identity (ID) labelling of 
positive and negative person image pairs 
for every camera-pair is prohibitively 
expensive, given that there are a quadratic number of camera pairs in a surveillance network.

It is no surprise that person re-id by 
{\em unsupervised learning} become
a focus in recent research. 
In this setting, per-camera-pair ID labelled
training data is no longer required 
\cite{wang2014unsupervised,kodirov2015dictionary,lisanti2015person,kodirov2016person,khan2016unsupervised,wang2016towards,ma2017person,ye2017dynamic,liu2017stepwise}. 
However, existing unsupervised learning re-id models are
significantly inferior in re-id accuracy. 
This is because, lacking cross-view pairwise ID labelled data deprives
a model's ability to learn strong discriminative
information. 
This nevertheless is critical for handling significant appearance change
across cameras. 

An alternative approach is to leverage jointly
{(1)} unlabelled data from a target domain which is freely available,
e.g. videos of thousands of people travelling through a camera view everyday in a public scene,
and 
{(2)} pairwise ID labelled datasets from independent source domains
\cite{want2018Transfer,peng2018joint,fan2017unsupervised,yu2017cross}. 
The main idea is to first learn
a ``view-invariant'' representation from ID labelled source data,
then adapt the pre-learned model to a target domain by using only unlabelled
target data. 
This approach makes an implicit assumption that,
the source and target domains share some common cross-view
characteristics so that a view-invariant representation can be estimated. This is not always true. 

In this work, we consider a {\em pure} unsupervised person re-id deep
learning problem. That is, no ID labelled training data are
assumed, neither cross-view 
nor within-view ID labelling.
Although this learning objective shares some modelling spirit 
with two recent domain
transfer models \cite{fan2017unsupervised,want2018Transfer}, both
those models do require {\em suitable} 
person ID labelled source domain training data, 
i.e. visually similar to the target domain.
Specifically, we consider unsupervised re-id model learning 
by jointly optimising unlabelled person tracklet data {\em
within-camera} view to be more discriminative and {\em cross-camera}
view to be more associative {\em end-to-end}. 

Our {\bf contributions} are:
We formulate a novel unsupervised person re-id deep learning method
using automatically generated person tracklets.
This avoids the need for
camera pairwise ID labelled training data,
i.e. {\em unsupervised tracklet re-id discriminative learning}.
Specifically, we propose a \textbf{\em Unsupervised Tracklet Association Learning} (UTAL) model with two key ideas: 
{\bf(1)} {\em Per-Camera Tracklet Discrimination Learning} that 
optimises ``local'' within-camera tracklet label discrimination.
It aims to facilitate cross-camera tracklet association given per-camera independently
created tracklet label spaces. 
{\bf(2)} {\em Cross-Camera Tracklet Association Learning} that 
\changed{optimises ``global'' cross-camera tracklet matching. It aims to find cross-view tracklet groupings that are most likely of the same person identities without ID label information.}
This is formulated as to jointly 
discriminate within-camera tracklet identity semantics
and self-discover
cross-camera tracklet pairwise matching 
in end-to-end deep learning. 
Critically, the proposed UTAL method does not assume 
any domain-specific knowledge
such as camera space-time topology and cross-camera ID overlap.
Therefore, it is scalable to arbitrary 
surveillance camera networks with unknown viewing conditions
and background clutters.

Extensive comparative experiments are conducted on
seven
existing benchmarking datasets 
(CUHK03 \cite{li2014deepreid}, 
Market-1501 \cite{zheng2015scalable}, 
DukeMTMC-ReID \cite{ristani2016performance,zheng2017unlabeled}, 
MSMT17 \cite{wei2018person},
iLIDS-VID \cite{wang2014person},
PRID2011 \cite{hirzer2011person},
MARS \cite{zheng2016mars})
and one newly introduced tracklet
person re-id dataset called DukeTracklet.
The results 
show the performance advantages and superiority of the proposed UTAL method 
over the state-of-the-art
unsupervised and domain adaptation person re-id models.

A preliminary version of this work
was reported in \cite{li2018Unsupervised}. 
Compared with the earlier study,
there are a few key differences introduced:
{\bf(i)} This study presents a more principled and scalable
unsupervised tracklet learning method 
that learns deep neural network re-id models 
directly from large scale raw tracklet data.
The method in \cite{li2018Unsupervised}
requires a separate preprocessing for domain-specific
spatio-temporal tracklet sampling
for reducing the tracklet ID duplication rate per camera view.  
This need for pre-sampling not only makes model learning more 
complex, not-end-to-end therefore suboptimal, but also loses a 
large number of tracklets with potential rich information useful
for more effective model learning.
{\bf(ii)} We propose in this study a new concept of soft tracklet labelling,
which aims to explore any inherent space-time visual correlation of the 
same person ID between unlabelled tracklets within each camera view. 
This is designed to better address the tracklet fragmentation 
problem through an end-to-end model optimisation mechanism.
\changed{It improves person tracking within individual camera views,}
which is lacking in
\cite{li2018Unsupervised}. 
{\bf(iii)} Unlike the earlier method, the current model self-discovers and exploits {\em explicit}
cross-camera tracklet association in terms of 
person ID, improving the capability of
re-id discriminative unsupervised learning
and leading to superior model performances.
{\bf(iv)} Besides creating a new tracklet person re-id dataset,
we further conduct more comprehensive evaluations 
and analyses 
for giving useful and significant insights.

\section{Related Work}
\label{sec:rel_work}

{\bf Person Re-Identification. }
Most existing person re-id models are built by {\em supervised} model learning
on a separate set of per-camera-pair ID labelled training data
\cite{li2014deepreid,chen2018person,li2017person,li2018harmonious,wei2018person,song2018mask,chang2018multi,shen2018deep}.
While having no class intersection,
the training and testing data are often assumed to be drawn from the same 
camera network (domain).
%
Their scalability 
is therefore significantly poor for 
realistic applications 
when no such large training sets are available
for every camera-pair in a test domain.
Human-in-the-loop re-id provides a means of reducing 
the overall amount of training label supervision by exploring 
the benefits of human-computer interaction
\cite{wang2016human,liu2013pop}.
%
But, it is still labour intensive and tedious.
Human labellers need to be deployed
repeatedly for conducting similar screen profiling operations
whenever a new target domain exhibits. 
It is therefore not scalable either.

Unsupervised model learning is an intuitive solution to 
avoiding the need of exhaustively collecting a large set
of labelled training data per application domain.
\changed{However, previous hand-crafted features-based 
unsupervised learning methods 
offer significantly inferior re-id matching performance \cite{farenzena2010person,ma2017person,
kodirov2015dictionary,kodirov2016person,khan2016unsupervised,ye2017dynamic,liu2017stepwise,
lisanti2015person,wang2014unsupervised,zhao2013unsupervised}},
when compared to the supervised learning models.
A trade-off between re-id model scalability and generalisation performance
can be achieved by semi-supervised learning
\cite{liu2014semi,wang2016towards}. 
But these models still assume
sufficiently large sized cross-view pairwise ID labelled data for model training.

There are attempts on unsupervised learning by
domain adaptation
\cite{want2018Transfer,fan2017unsupervised,peng2018joint,yu2017cross,zhu2017unpaired,deng2018image,zhong2018generalizing}.
The idea is to exploit the knowledge of labelled data in
``related'' source domains through model adaptation on the unlabelled target
domain data.
One straightforward approach is
to convert the source ID labelled training data
into the target domain by appearance mimicry.
This enables to train a model using the domain style transformed source training data
via supervised learning
\cite{zhu2017unpaired,deng2018image}.
Alternative techniques include 
semantic attribute knowledge transfer \changed{\cite{peng2016unsupervised,peng2018joint,want2018Transfer},
space-time pattern transfer \cite{lv2018unsupervised},
virtual ID synthesis \cite{bak2018domain},}
and progressive 
adaptation
\cite{fan2017unsupervised,yu2017cross}.
%
While these 
models perform better than the
earlier generation of methods (Tables~\ref{tab:img_SOTA} and~\ref{tab:vide_SOTA}),
they require similar data distributions and viewing
conditions between the labelled source domain and the unlabelled
target domain. This restricts their scalability to arbitrarily
diverse (and unknown) target domains in large scale deployments.

Unlike all existing unsupervised learning re-id methods,
the proposed tracklet association method in this work enables unsupervised
re-id deep learning from scratch at end-to-end. 
This is more scalable and general.
Because there is no assumption on either the 
scene characteristic similarity between source and target domains, 
or the complexity of handling 
ID label knowledge transfer.
Our method directly learns to discover the re-id discriminative knowledge
from {\em unlabelled} tracklet data automatically generated.

Moreover, the proposed method does not assume any overlap of person ID classes across camera views or other domain-specific information.
It is therefore scalable to 
the scenarios without any knowledge about
camera space-time topology \cite{lv2018unsupervised}.
%
Unlike the existing unsupervised learning method relying on extra hand-crafted features,
our model learns tracklet based re-id discriminative features from an
end-to-end deep learning process. 
To our best knowledge, this is the {\em first} attempt at unsupervised
tracklet association based person re-id deep learning model without
relying on any ID labelled training video or imagery data.

\noindent{\bf Multi-Task Learning in Neural Networks. }
Multi-task learning (MTL) is a machine learning strategy
that learns several related tasks simultaneously for 
their mutual benefits \cite{argyriou2007multi}. 
A good MTL survey with focus on neural networks
is provided in \cite{caruana1997multitask}.  
Deep CNNs are well suited for 
performing MTL. As they are inherently designed to learn joint 
feature representations subject to multiple label objectives
concurrently in multi-branch architectures.
Joint learning of multiple related tasks 
has been proven to be effective in solving computer vision problems
\cite{dong2017multi,zhang2016learning}.
%

In contrast to all the existing methods 
aiming for supervised learning problems,
the proposed UTAL method exploits differently the MTL principle to solve an unsupervised learning task.
Critically, our method is uniquely designed to 
explore the potential of MTL
in correlating the underlying {\em group} level semantic relationships
between different individual learning tasks\footnote{In the unsupervised tracklet person re-id context, a group corresponds to
a set of categorical labels 
each associated with an individual person tracklet 
drawn from a specific camera view.}.
This dramatically differs from existing MTL based methods
focusing on mining the shared knowledge among tasks
at the sample level.
Critically, it avoids the simultaneous labelling of multi-tasks
on each training sample.
%
%
Sample-wise multi-task labels are not available 
in the unsupervised tracklet re-id problem.

Besides, unsupervised 
tracklet labels in each task (camera view)
are {\em noisy}.
As they are obtained without manual verification. 
Hence, the proposed UTAL model is in effect performing
{\em weakly supervised multi-task learning} 
with noisy per-task labels.
This makes our method fundamentally different from existing MTL approaches 
that are only interested in discovering discriminative cross-task common representations
by {\em strongly supervised learning} of clean and exhaustive sample-wise multi-task labelling.

\noindent {\bf Unsupervised Deep Learning. }
Unsupervised learning of visual data
is a long standing research problem starting from
the auto-encoder models \cite{olshausen1997sparse}
or earlier. 
Recently, this problem has regained attention
in deep learning. 
One common approach is by incorporating with data clustering 
\cite{xie2016unsupervised} 
that jointly learns deep feature representations and image clusters. 
Alternative unsupervised learning techniques include 
formulating generative models 
\cite{zhu2017unpaired},
devising a loss function that preserves information flowing 
\cite{bojanowski2017unsupervised}
or discriminates instance classes \cite{wu2018unsupervised},
exploiting the object tracking continuity cue \cite{wang2015unsupervised}
in unlabelled videos, and so forth.

As opposite to all these methods focusing on uni-domain data distributions,
our method is designed particularly to 
\changed{learn visual data} sampled from 
different camera domains with unconstrained viewing settings.
Conceptually, the proposed idea of soft tracklet labels
is related to data clustering. As the per-camera affinity matrix 
used for soft label inference is related to 
the underlying data cluster structure. 
However, our affinity based method has the unique merits of 
avoiding per-domain hard clustering and having fewer parameters 
to tune (e.g. the per-camera cluster number).

\begin{figure*}[t]
	\centering
	\includegraphics[width=0.95\textwidth]{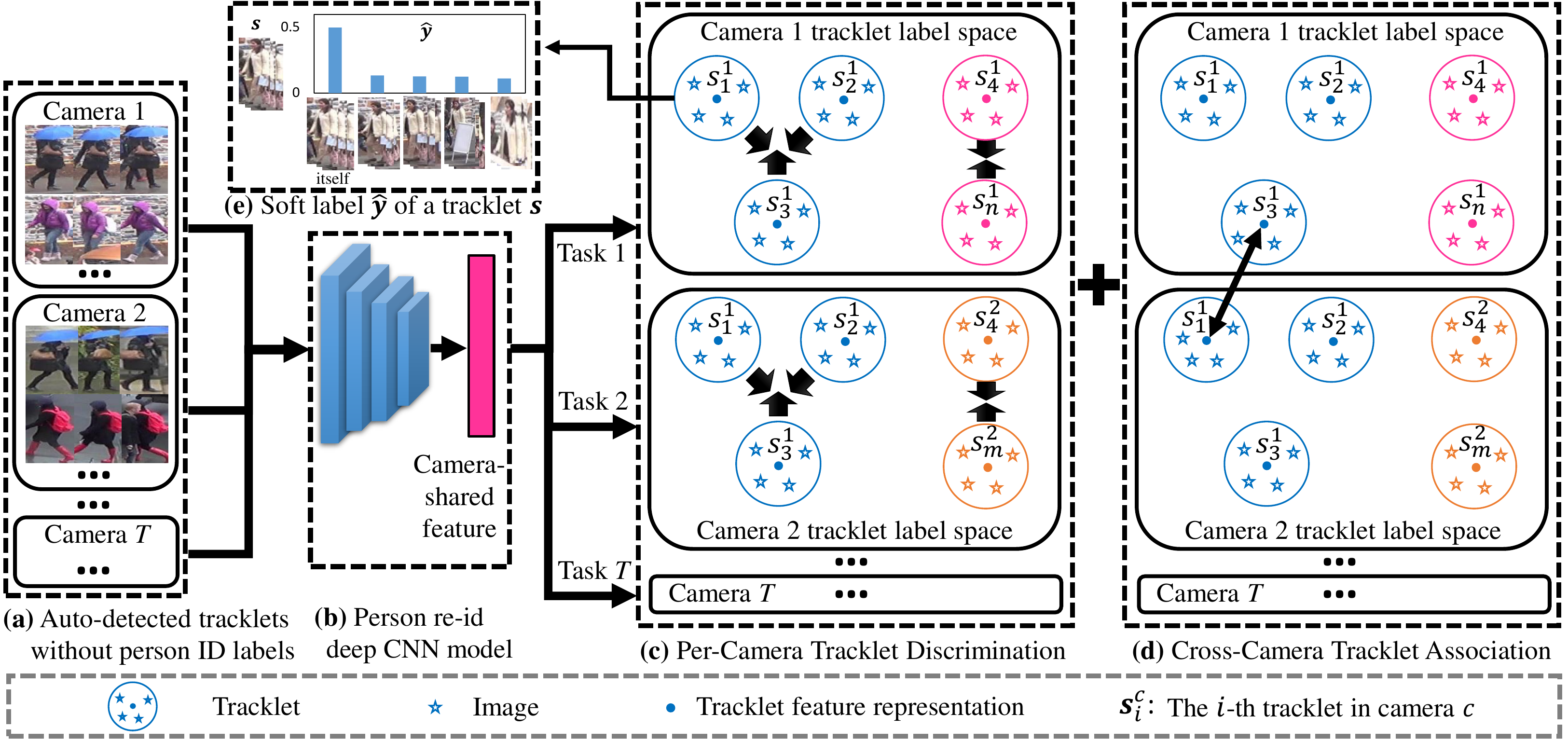}
	\vskip -0.3cm
	\caption{
		An overview of the proposed {\em Unsupervised Tracklet Association Learning} 
		(UTAL) person re-identification model. 
		The UTAL takes as input {\bf (a)} auto-detected tracklets from all the camera views 
		without any person ID class labelling either within-camera or cross-camera. 
		The objective is to derive {\bf (b)} a person re-id discriminative feature representation model
		by unsupervised learning.
		To this end, we formulate the UTAL model 
		for simultaneous {\bf(c)} Per-Camera Tracklet Discrimination (PCTD) learning and
		{\bf(d)} Cross-Camera Tracklet Association (CCTA) learning
		in an end-to-end neural network architecture.
		The PCTD aims to derive the ``local'' discrimination of per-camera tracklets
		in the respective tracklet label space 
		\changed{(represented by soft labels {\bf(e)})}
		by a multi-task inference process (one task for a specific camera view),
		whilst the CCTA to learn the ``global'' cross-camera tracklet association 
		across independently formed tracklet label spaces.
		In UTAL design, the PCTD and CCTA jointly learn to optimise a re-id model for maximising their 
		complementary contributions and advantages in a synergistic interaction
		and integration.
		Best viewed in colour.
	}
	\label{fig:pipeline}
	\vspace{-0.3cm}
\end{figure*}

\section{Method Formulation} \label{sec:method}

To overcome the limitation of supervised model learning algorithms
for exhaustive within-camera and cross-camera ID labelling, 
we propose a novel {Unsupervised Tracklet Association Learning} (UTAL)
method to person re-id in videos (or multi-shot images in general).
This is achieved by
exploiting person {\em tracklet labelling} obtained from 
existing trackers\footnote{Although object tracklets can be generated 
by any independent single-camera-view multi-object tracking (MOT) models widely available currently, 
a conventional MOT model 
is {\em not} end-to-end optimised for cross-camera tracklet association.}
{\em without}
any ID labelling either cross-camera or within-camera.
The UTAL learns a person re-id model end-to-end 
therefore
benefiting from joint overall model optimisation in deep learning.
In the follows, we first present unsupervised per-camera
tracklet labelling (Sec. \ref{sec:tracklet_labelling}),
then describe our model design for 
within-camera and cross-camera tracklet association
by joint unsupervised deep learning (Sec. \ref{sec:tracklet_asso_learning}).

\subsection{Unsupervised Per-Camera Tracklet Formation} 
\label{sec:tracklet_labelling}

Given a large quantity of video data captured by \changed{disjoint} surveillance cameras,
we first deploy the off-the-shelf
pedestrian detection and tracking models
\cite{liu2016ssd,zhang2016far,leal2015motchallenge}
to automatically extract person tracklets. 
We then 
annotate each tracklet \changed{$\bm{S}$} with a unique class (one-hot) label 
\changed{$y$}
in an {\em unsupervised} and {\em camera-independent} manner.
This does not involve any {\em manual} ID verification on tracklets.
By applying this tracklet labelling method 
in each camera view separately, 
we can obtain an {\em independent} set of labelled tracklets $\{\bm{S}_i, y_i\}$
per camera,
where each tracklet $\bm{S}$ contains a varying number of 
person bounding box images $\bm{I}$ as $\bm{S} = \{\bm{I}_1, \bm{I}_2, \cdots \}$.

\noindent \textbf{Challenges. } 
To effectively learn a person re-id model from such automatically labelled tracklet training data,
we need to deal with two modelling challenges
centred around the supervision of person ID class labels: 
(1) Due to frequent trajectory fragmentation,
multiple tracklets (unknown due to no manual verification) 
are often generated during the appearing period of a person under one camera view.
However, they are unsupervisedly assigned with different one-hot categorical labels.
This may significantly mislead the discriminative learning process of a re-id model.
(2) There are no access to 
positive and negative pairwise ID correspondence 
between tracklet labels across \changed{disjoint} camera views.
Lacking cross-camera person ID supervision 
underpins one of the key challenges in unsupervised tracklet person re-id.

\subsection{Unsupervised Tracklet Association}
\label{sec:tracklet_asso_learning}

Given per-camera independent tracklets  
$\{\bm{S}_i, y_i\}$, 
we 
explore {\em tracklet label re-id discriminative learning}
without person ID labels in a 
deep learning classification framework.
We formulate an \textbf{\em Unsupervised Tracklet Association Learning} (UTAL) method, with
the overall architecture design 
illustrated
in Fig. \ref{fig:pipeline}.
The UTAL contains two model components:
{\bf(I)} {\em Per-Camera Tracklet Discrimination} (PCTD) learning 
\changed{for optimising ``local'' within-camera tracklet label discrimination.
This facilitates ``global'' cross-camera tracklet association,
given {\em independent} tracklet label spaces in different camera views.}
{\bf(II)} {\em Cross-Camera Tracklet Association} (CCTA) learning 
for discovering ``global'' cross-camera tracklet identity matching
without ID labels. 

\begin{figure} [t] 
	\centering
	\includegraphics[width=\columnwidth]{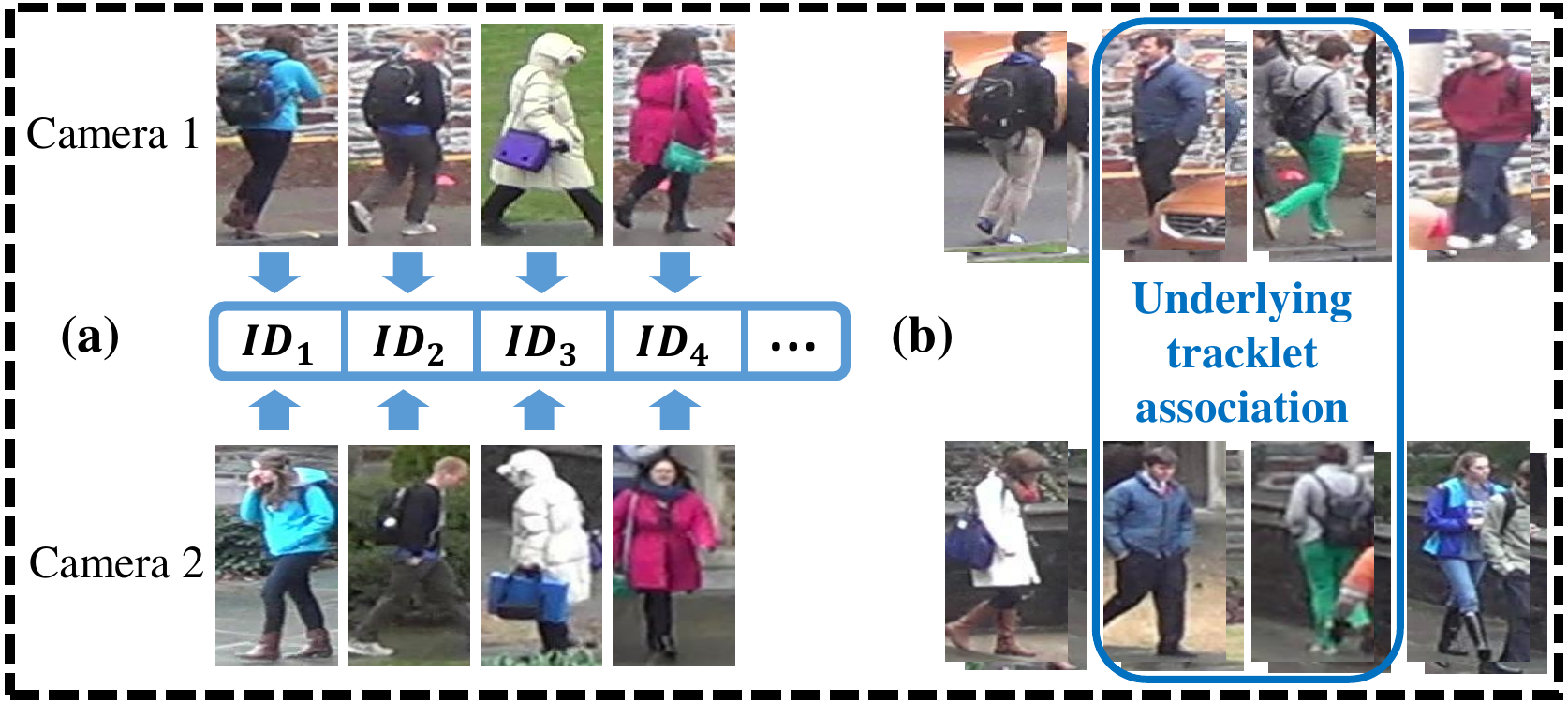}
	\vskip -0.3cm
	\caption{Comparing
		{\bf (a)} Fine-grained {\em explicit instance-level}
          cross-camera ID labelled image pairs for supervised person
          re-id model learning and 
		{\bf (b)} Coarse-grained {\em latent group-level} cross-camera
                tracklet (a multi-shot group) label correlation for ID label-free
                (unsupervised) person re-id learning using the proposed UTAL method. 
	}
	\vspace{-0.3cm}
	\label{fig:label_info}
\end{figure}

For accurate cross-camera tracklet association,
it is important to formulate a robust image feature representation to
characterise the person appearance of each tracklet.
%
However, it is sub-optimal to achieve ``local'' per-camera tracklet
discriminative learning using only per-camera independent tracklet labels
without ``global'' cross-camera tracklet correlations. We therefore propose to
optimise jointly both PCTD and CCTA.
The two components integrate as a whole in a single deep learning 
architecture, learn jointly and mutually benefit each other
in incremental end-to-end model optimisation.
Our overall idea for unsupervised learning of tracklet person re-id is  
to maximise coarse-grained {\em latent group-level}
cross-camera tracklet association.
This is based on exploring an 
notion of tracklet set correlation learning
(Fig. \ref{fig:label_info}(b)).
It differs significantly from 
supervised re-id learning 
that 
relies heavily on 
the fine-grained {\em explicit instance-level} cross-camera
ID pairwise supervision 
(Fig. \ref{fig:label_info}(a)).

\subsubsection{Per-Camera Tracklet Discrimination Learning}
In PCTD learning, we treat each individual camera view separately. That is,
optimising per-camera labelled tracklet discrimination as a
classification task with the unsupervised per-camera tracklet labels 
(not person ID labels) (Fig. \ref{fig:pipeline}(a)). 
Given a surveillance network with $T$ cameras, 
we hence have a total of $T$ different tracklet classification tasks
each corresponding to a specific camera view.

Importantly, we further formulate these $T$ classification tasks
in a multi-branch network architecture design. 
All the tasks share the {\em same} feature representation space (Fig. \ref{fig:pipeline}(b)) 
whilst enjoying an individual classification branch (Fig. \ref{fig:pipeline}(c)).
%
This is a multi-task learning \cite{caruana1997multitask}.
%

Formally, we assume $M_t$ different tracklet labels $\{y\}$ 
with the training tracklet image frames 
$\{\bm{I}\}$ 
from a camera view $t \in \{1, \cdots,T\}$ (Sec. \ref{sec:tracklet_labelling}).
We adopt the softmax Cross-Entropy (CE) loss function
to optimise the corresponding classification task (the $t$-th branch).
The CE loss on a training image frame $(\bm{I}, y)$ is computed as: 
\begin{equation}
\mathcal{L}_\text{ce}=
{-}
\sum_{j=1}^{M_t} \mathbbm{1} (j=y)
\cdot
{\log}\Big(\frac{\exp({\bm{W}_{j}^{\top} {\bm x}})}
{\sum_{k=1}^{M_t}\exp({\bm{W}_{k}^{\top} {\bm x}})}\Big)
\label{eq:CE_loss}
\end{equation}
where
$\bm{x}$ specifies the feature vector of $\bm{I}$
extracted from the {\em task-shared} representation space 
and 
$\bm{W}_{y}$ the $y$-th class prediction parameters. 
$\mathbbm{1} (\cdot)$ denotes an indicator function
that returns $1$/$0$ 
for true/false arguments.
Given a training mini-batch, 
we compute the CE loss for each such training sample
with the respective tracklet label space
and utilise their average to form the PCTD learning objective as:
\begin{equation}
\mathcal{L}_\text{pctd}= \frac{1}{N_\text{bs}} \sum_{t=1}^{T}
\mathcal{L}_\text{ce}^t
\label{eq:PCTD_loss}
\end{equation}
where $\mathcal{L}_\text{ce}^t$ denotes the CE loss of all training tracklet frames
from the $t$-th camera, 
and $N_\text{bs}$ specifies the batch size.

Recall that, one of the main challenges in unsupervised tracklet re-id learning
arises from within-camera trajectory fragmentation.
This
causes the tracklet ID duplication issue, i.e. 
the same-ID tracklets 
are assigned with distinct labels.
By treating every single tracklet label as a unique class (Eq \eqref{eq:PCTD_loss}),
misleading supervision can be resulted 
potentially hampering the model learning performance.

\noindent {\bf Soft Labelling. }
For gaining learning 
robustness against unconstrained trajectory fragmentation,
we exploit the pairwise appearance affinity (similarity) information between within-camera tracklets.
To this end, we propose {\em soft tracklet labels}
to replace the {\em hard} counterpart (one-hot labels).
This scheme uniquely 
takes into account the 
underlying ID correlation between tracklets in the PCTD learning 
(Eq \eqref{eq:PCTD_loss}).
It is based on the intuition that,
multiple fragmented tracklets of the same person are more likely 
to share higher visual affinity with each other 
than those describing different people. 
Therefore, using tracklet labels involving 
the appearance affinity (i.e. soft labels) 
means imposing person ID relevant information 
into model training, from the manifold structure learning perspective \cite{belkin2006manifold}.

Formally, we start the computation of soft tracklet labels 
by
constructing an affinity matrix of person appearance 
$\mathcal{A}^t \in\mathbb{R}^{M_{t} \times M_{t}}$ 
on all $M_{t}$ tracklets for each camera $t \in \{1, \cdots,T\}$, 
where each element
$\mathcal{A}^t(i,j)$ specifies the visual appearance similarity between the tracklets $i$ and $j$.
This requires
a tracklet feature representation space.
We achieve this by cumulatively updating 
an external feature vector $\bm{s}$ for every single tracklet $\bm{S}$
with the image features of the constituent frames
in a batch-wise manner.

More specifically, given a mini-batch 
including $n_i^t$ image frames from 
the $i$-th tracklet $\bm{S}_i^t$ of $t$-th camera view,
the corresponding tracklet representation $\bm{s}_i^t$ is 
progressively updated across the training iterations as:
\begin{equation}
	\bm{s}_i^t = \frac{1}{1+\alpha}\Big[\bm{s}_i^t + \alpha \Big( \frac{1}{n_i^t} \sum_{k=1}^{n_i^t} \bm{x}_k \Big)\Big]
	\label{eq:tracklet_feat}
\end{equation}
where $\bm{x}_k$ is the feature vector of the $k$-th in-batch image frame of $\bm{S}_i^t$,
extracted by the up-to-date model.
The learning rate parameter $\alpha$ controls how fast $\bm{s}_i^t$ updates. 
This method avoids the need of forwarding all tracklet data in each iteration
therefore computationally efficient and scalable.

Given the tracklet feature representations,
we subsequently 
sparsify the affinity matrix as:
\begin{equation}
	\mathcal{A}^t(i,j) = \left\{
	\begin{array}{ll}
		\exp(-\frac{{\left\| \bm{s}_i^t-\bm{s}_j^t \right\|}^2_2}{\sigma^2}), 
		& \text{if} \;\; \bm{s}_j^t \in {\mathcal{N}(\bm{s}_i^t)}\\
		0, & \text{otherwise}
	\end{array}
	\right.
	\label{eq:sparse}
\end{equation}
where $\mathcal{N}(\bm{s}_i^t)$ are the $K$ nearest neighbours (NN) of $\bm{s}_i^t$ 
defined by the Euclidean distance in the feature space. 
Using the sparse NN idea on the affinity structure is for suppressing the distracting effect of visually similar tracklets from unmatched ID classes.
\changed{Computationally, each $\mathcal{A}$ has a quadratic complexity,
  but only to the number of {\em per-camera}
  tracklet and linear to the total number of cameras, rather than
  quadratic to all tracklets from all the cameras.
The use of sparse similarity matrices significantly reduces the memory demand.  
}

To incorporate the local density structure \cite{zelnik2005self},
we deploy a neighbourhood structure-aware scale defined as:
\begin{equation}
	\sigma^2= \frac{1}{M_{t} \cdot K} 
	\sum_{i=1}^{M_t} \sum_{j=1}^{K}
	{\left\| \bm{s}_i^t-\bm{s}_j^t \right\|}^2_2, \;\; s.t. \;\; \bm{s}_j^t \in \mathcal{N}(\bm{s}_i^t)
	\label{eq:sigma}
\end{equation}
	
Based on the estimated neighbourhood structures, 
we finally compute the soft label \changed{(Fig. \ref{fig:pipeline}(e))} 
for each tracklet $\bm{S}_i^t$ as 
the $L_1$ normalised affinity measurement:
\begin{equation}
	\hat{\bm{y}}_i^t = \frac{\mathcal{A}(i,1:M_t)}{\sum_{j=1}^{M_t}\Big(\mathcal{A}(i,j)\Big)}
	\label{eq:soft_label}
\end{equation}

Given the proposed soft tracklet labels, 
the CE loss function (Eq \eqref{eq:PCTD_loss}) is then reformulated as: 
\begin{equation}
	\mathcal{L}_\text{sce} = 
	{-} \sum_{j=1}^{M_t} 
	\hat{\bm{y}}_i^t(j) 
	\cdot
	{\log}\Big(\frac{\exp({\bm{W}_{j}^{\top} {\bm x}})}
	{\sum_{k=1}^{M_t}\exp({\bm{W}_{k}^{\top} {\bm x}})}\Big)
	\label{eq:soft_CE_loss}
\end{equation}
We accordingly update the PCTD learning loss (Eq \eqref{eq:PCTD_loss}) as:
\begin{equation}
\mathcal{L}_\text{pctd}= \frac{1}{N_\text{bs}} \sum_{t=1}^{T}
\mathcal{L}_\text{sce}^t.
\label{eq:PCTD_loss_final}
\end{equation}

\noindent \textbf{Remarks. } 
In PCTD, the objective function
(Eq \eqref{eq:PCTD_loss_final}) optimises by supervised learning
person tracklet discrimination {\em within} each camera view alone. 
It does not explicitly consider
supervision in {\em cross-camera} tracklet association. 
Interestingly, when jointly learning all the per-camera tracklet
discrimination tasks together, the learned representation model is
{\em implicitly} and {\em collectively} cross-view
tracklet discriminative in a latent manner.
This is due to the existence of cross-camera tracklet
ID class correlation.
That being said,
the shared feature representation is optimised to
be tracklet discriminative {\em concurrently} for 
all camera views, 
latently expanding model discriminative learning from per-camera ({\em locally})
to cross-camera ({\em globally}).

Apart from multi-camera multi-task learning, 
we exploit the idea of soft tracklet labels
to further improve the model ID discrimination learning capability.
This is for better robustness against trajectory fragmentation.
Fundamentally, this is an indirect strategy of refining fragmented tracklets.
It is based on the visual appearance affinity without the need of explicitly
stitching tracklets.
The intuition is that, the tracklets of the same person are possible to 
be assigned with more similar soft labels (i.e. signatures).
Consequently, this renders the unsupervised tracklet labels closer to
supervised ID class labels in terms of discrimination power,
therefore helping re-id model optimisation. 

In equation formulation, our PCTD objective is related to 
the Knowledge Distillation (KD) technique \cite{hinton2015distilling}.
KD also utilises soft class probability labels
inferred by an independent teacher model. 
Nevertheless, our method conceptually differs from KD,
since we primarily aim to unveil the hidden fine-grained discriminative information
of the same class (ID) distributed across unconstrained tracklet fragments,
Besides, our model retains the KD's merit of modelling
the inter-class similarity geometric manifold information.
Also, our method has no need for learning a heavy source knowledge teacher model,
therefore, computationally more efficient.
We will evaluate 
the PCTD model design 
(Table \ref{tab:PCJL}).

\subsubsection{Cross-Camera Tracklet Association Learning }
The PCTD 
achieves somewhat global (all the camera views) tracklet discrimination capability {\em implicitly}. 
But the resulting representation remains sub-optimal, due to the lack of {\em
  explicitly} optimising cross-camera tracklet association at the
fine-grained instance level.
It is non-trivial to impose cross-camera re-id
discriminative learning constraints
at the absence of ID labels. 
To address this problem,
we introduce a Cross-Camera Tracklet Association (CCTA) learning algorithm 
for enabling tracklet association between cameras (Fig. \ref{fig:pipeline}(d)). 
Conceptually, the CCTA is based on 
{\em adaptively and incrementally self-discovering cross-view tracklet association}
in the multi-task camera-shared feature space (Fig. \ref{fig:pipeline}(b)). 

Specifically, we ground the CCTA learning 
on cross-camera nearest neighbourhoods.	
In re-id, the vast majority of cross-camera tracklet pairs
are negative associations from unmatched ID classes.
They provide no
desired information about how a person's appearance
varies under different camera viewing conditions. 
The key for designing an informative CCTA loss is therefore
to self-discover the cross-camera positive {\em matching} pairs.
This requires to search similar samples (i.e. neighbours) 
which however is a challenging task because:
(1) Tracklet feature representations $\bm{s}$ can be unreliable and error-prone
due to the lack of cross-camera pair supervision (hence a catch-22 problem).
(2) False positive pairs may easily propagate the erroneous supervision 
cumulatively through the learning process, guiding the optimisation 
towards poor local optima;
Deep neural networks possess the capacity to 
fit any supervision labels \cite{zhang2016understanding}.

To overcome these challenges, 
we introduce a model matureness adaptive matching pair search mechanism.
It progressively finds an {\em increasing} number of plausible true matches across cameras
as a reliable basis for the CCTA loss formulation.
In particular, 
in each training epoch, 
we first retrieve the reciprocal {cross-camera} nearest neighbour $\mathcal{R}(\bm{s}_i^t)$
for each tracklet $\bm{s}_i^t$.
The $\mathcal{R}$ is obtained based on the mutual nearest neighbour notion \cite{qin2011hello}.
Formally, let $\mathcal{N}^1(\bm{s}_i^t)$ be 
the {cross-camera} NN
of $\bm{s}_i^t$.
The $\mathcal{R}(\bm{s}_i^t)$ is then defined as:
\begin{equation}
\mathcal{R}(\bm{s}_i^t) = 
\{\bm{s} | \bm{s} \in \mathcal{N}^1(\bm{s}_i^t) \;\; \&\& \;\; \bm{s}_i^t \in \mathcal{N}^1(\bm{s}) \}
\label{eq:rep_knn}
\end{equation}

Given such self-discovered cross-camera matching pairs,
we then formulate a CCTA objective loss for a tracklet $\bm{s}_i^t$ as:
%
\begin{equation}\label{eq:CCTA}
\mathcal{L}_\text{ccta}
= {\sum_{\bm{s} \in {\mathcal{R}(\bm{s}_i^t)}} \parallel \bm{s}^{t}_{i} - \bm{s} \parallel_{2}}
\end{equation}

With Eq~\eqref{eq:CCTA}, we impose a cross-camera discriminative learning constraint
by encouraging the model to pull the neighbour tracklets in ${\mathcal{R}_i^t}$
close to $\bm{s}_i^t$.
This CCTA loss applies only to those tracklets $\bm{s}$
with cross-camera matches, i.e. $\mathcal{R}(\bm{s})$ is non-empty,
so that it is model matureness adaptive.
As the training proceeds, the model is supposed to become more mature,
leading to more cross-camera tracklet matches discovered.
We will evaluate the CCTA loss in Sec. \ref{sec:exp_component}.

\noindent \textbf{Remarks. }
Under the mutual nearest neighbour constraint,  
$\mathcal{R}(\bm{s}_i^t)$ are considered to be
more strictly similar 
to $\bm{s}_i^t$ than 
the conventional nearest neighbours $\mathcal{N}(\bm{s}_i^t)$. 
With uncontrolled viewing condition variations across cameras, 
matching tracklets with dramatic appearance changes may be excluded from the $\mathcal{R}(\bm{s}_i^t)$ particularly 
in the beginning, 
representing a
conservative search strategy.
This is designed so to minimise the negative effect 
of error propagation 
from false matching pairs.
More matching pairs are likely unveiled
as the training proceeds.
Many previously missing
pairs can be gradually discovered when the model becomes more mature and discriminative.
Intuitively, easy matching pairs are found before hard ones.
Hence, the CCTA loss is in a curriculum leaning spirit 
\cite{bengio2009curriculum}.
In Eq \eqref{eq:CCTA} we consider only the 
positive pairs whilst ignoring the negative matches.
This is conceptually analogue to the formulation of Canonical Correlation Analysis (CCA)
\cite{hardoon2004canonical}, and results in
a simpler objective function without
the need to tune a margin hyper-parameter as required by
the ranking losses \cite{schroff2015facenet}.

\subsubsection{Joint Unsupervised Tracklet Association Learning}\label{sec:joint-unsupervised-tracklet-association-learning}
By combining the CCTA and PCTD learning constraints, we obtain the final 
model objective loss function of UTAL as: 
\begin{equation}\label{eq:loss_UTAL}
\mathcal{L}_\text{utal}= 
\mathcal{L}_\text{pctd} + \lambda \mathcal{L}_\text{ccta}
\end{equation}
where $\lambda$ is a balance weight. 
Note that $\mathcal{L}_\text{pctd}$ is an average loss term at the individual tracklet image level
whilst $\mathcal{L}_\text{ccta}$ at the tracklet group (set) level.
Both are derived from the same mini-batch of training data concurrently.

\noindent {\bf Remarks. }
By design, the CCTA enhances model representation learning. 
It imposes discriminative constraints derived from self-discovered cross-camera tracklet association.
This is based on the PCTD learning of unsupervised and 
per-camera independent tracklet label spaces.
%
With more discriminative representation
in the subsequent training iterations, 
the PCTD is then able to deploy more accurate soft tracklet labels. 
This in turn facilitates 
not only the following representation learning 
of per-camera tracklets, but also the discovery of higher quality 
and more informative cross-camera tracklet matching pairs.
%
In doing so, the two learning components 
integrate seamlessly and optimise 
a person re-id model concurrently
in an end-to-end batch-wise learning process.
Consequently, the overall UTAL method formulates a
benefit-each-other closed-loop model design.
This eventually leads to 
cumulative and complementary advantages throughout training.

\subsubsection{Model Training and Testing}
\noindent \textbf{Model Training. }
To minimise the negative effect of inaccurate cross-camera tracklet matching pairs,
we deploy the CCTA loss {\em only} during the second half training process. 
Specifically,
UTAL begins the model training with the 
soft tracklet label based PCTD loss (Eq \eqref{eq:PCTD_loss_final})
for the first half epochs.
We then deploy the full UTAL loss (Eq \eqref{eq:loss_UTAL})
for the remaining epochs.
To improve the training efficiency, 
we update the per-camera tracklet soft labels (Eq \eqref{eq:soft_label})
and cross-camera tracklet matches (Eq \eqref{eq:rep_knn})
per epoch.
These updates progressively enhance the re-id discrimination power of the UTAL objective throughout training, as we will show in our model component analysis
and diagnosis in Sec. \ref{sec:exp_component}.

\noindent \textbf{Model Testing. }
Once a deep person re-id model is trained by the UTAL unsupervised learning method,
we deploy the 
camera-shared feature representations (Fig. \ref{fig:pipeline}(b)) 
for re-id matching under the Euclidean distance metric.

\section{Experiments}
\label{sec:Exp}

\subsection{Experimental Setting}
{\bf Datasets. }
To evaluate the proposed UTAL model, we tested both video
(iLIDS-VID \cite{wang2014person}, 
PRID2011 \cite{hirzer2011person},
MARS \cite{zheng2016mars}) 
and 
image
(CUHK03 \cite{li2014deepreid},
Market-1501 \cite{zheng2015scalable}, 
DukeMTMC-ReID \cite{ristani2016performance,zheng2017unlabeled}, 
MSMT17 \cite{wei2018person}) 
person re-id 
datasets. 
In previous studies, these two sets of benchmarks 
were usually evaluated {\em separately}.
We consider both sets because recent large image re-id datasets were typically constructed 
by sampling person bounding boxes from videos,
so they share
similar characteristics as the video re-id datasets.
We adopted the standard  
test protocols as summarised in Table \ref{tab:dataset_stats}.

\begin{figure}[t]
	\centering
	\includegraphics[width=\linewidth]{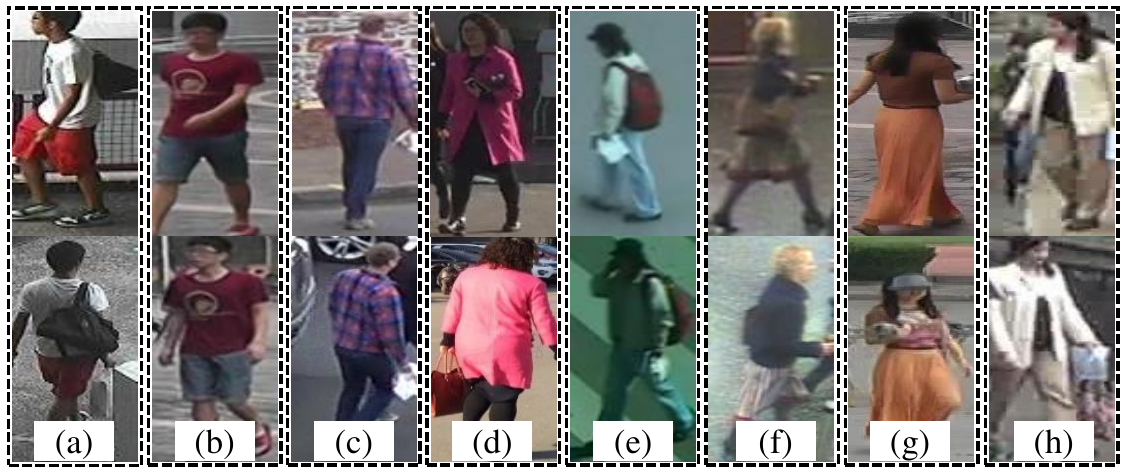}
	\vskip -0.4cm
	\caption{Example cross-camera matching image/tracklet pairs from 
		(a) CUHK03, (b) Market-1501, (c) DukeMTMC-ReID, (d) MSMT17, 
		(e) PRID2011, (f) iLIDS-VID, (g) MARS, (h) \changed{DukeMTMC-SI-Tracklet}.
	}
	\label{fig:dataset_img}
	\vspace{-0.3cm}
\end{figure}

\begin{table}[h] 
	\centering
	\setlength{\tabcolsep}{0.09cm}
	\caption{
		Dataset statistics and
		evaluation setting.
	}
	\vskip -0.3cm
	\begin{tabular}
		{c||c|c|c|c|c}
		\hline 
		Dataset  & 
		{\# ID} & 
		{\# Train } & 
		{\# Test} &
		{\# Image} & 
		{\# Tracklet} \\ \hline \hline %
		iLIDS-VID \cite{wang2014person}		& 300 & 150 & 150 & 43,800 & 600 			\\
		PRID2011 \cite{hirzer2011person}		& 178 & 89 & 89 & 38,466 & 354 				\\
		MARS \cite{zheng2016mars} 			& 1,261 & 625 & 636 & 1,191,003 & 20,478 	\\
		\em \changed{DukeMTMC-SI-Tracklet}		& 1,788 & 702 & 1,086 & 833,984 & 12,647 	\\
		\hline
		CUHK03 \cite{li2014deepreid}		& 1,467 & 767 & 700 & 14,097 & 0 		\\
		Market-1501 \cite{zheng2015scalable}		& 1,501& 751 & 750  & 32,668 & 0 		\\
		DukeMTMC-ReID \cite{zheng2017unlabeled}		& 1,812 & 702 & 1,110 & 36,411 & 0		\\
		MSMT17\cite{wei2018person} 				& 4,101 & 1,041 & 3,060 & 126,441 & 0 	\\
		\hline
	\end{tabular}
	\label{tab:dataset_stats}
	\vspace{-0.2cm}
\end{table}

\begin{table*} 
	\centering
	\setlength{\tabcolsep}{0.48cm}
	\caption{Unsupervised person re-id on image based datasets.
		$^*$: Benefited from extra labelled auxiliary training data. 
		``-'': No reported result.}
	\vskip -0.3cm
	\label{tab:img_SOTA}
	\begin{tabular}
		{l||c|c||c|c||c|c||c|c}
		\hline
		{Dataset}						
		& \multicolumn{2}	{c||}{CUHK03 \cite{li2014deepreid}}
		& \multicolumn{2}{c||}{Market-1501 \cite{zheng2015scalable}}                    	
		& \multicolumn{2}{c||}{DukeMTMC-ReID \cite{zheng2017unlabeled}}    						
		& \multicolumn{2}	{c}{MSMT17 \cite{wei2018person}}
		\\ \hline 
		Metric (\%) 
		& Rank-1 & mAP & Rank-1	& mAP & Rank-1 & mAP & Rank-1 & mAP
		\\ \hline \hline
		Dic \cite{kodirov2015dictionary}	& 36.5	& - 		 & 50.2 	& 22.7		& -  	& - 		& -   	& -		\\
		ISR \cite{lisanti2015person}		& 38.5   & -		 & 40.3 	& 14.3		& - 		& -         & -   	& -		\\
		RKSL \cite{wang2016towards}		& 34.8   & -		 & 34.0 	& 11.0		& -   	& - 		& -   	& -		\\
		\hline
		SAE$^*$ \cite{lee2008sparse}			& 30.5   & - 		 & 42.4  	& 16.2   		& -   	& - 		& -   	& -		\\
		JSTL$^*$ \cite{xiao2016learning}		& 33.2   & - 		 & 44.7 	& 18.4          & -   	& - 		& -   	& -		\\
		AML$^*$ \cite{ye2007adaptive}			& 31.4   & -		 & 44.7 	& 18.4   		& -   	& - 		& -   	& -		\\
		UsNCA$^*$ \cite{qin2015unsupervised}	& 29.6   & -		 & 45.2  	& 18.9   		& -   	& - 		& -   	& -		\\
		CAMEL$^*$ \cite{yu2017cross}		& 39.4 	& -	 	 & 54.5 	& 26.3          & -   	& - 		& -   	& -		\\
		PUL$^*$ \cite{fan2017unsupervised}	& -   	& -         & 44.7	& 20.1        	& 30.4 	& 16.4 	& -   	& -		\\
		TJ-AIDL$^*$ \cite{want2018Transfer}	& -  & - & 58.2 & 26.5 & 44.3 & 23.0 & - & - 
		\\
		CycleGAN$^*$ \cite{zhu2017unpaired} 
		& - & - & 48.1 & 20.7 & 38.5 & 19.9 & - & - 
		\\
		SPGAN$^*$ \cite{deng2018image} 
		& - & - & 51.5 & 22.8 & 41.1 & 22.3 & - & - 
		\\
		SPGAN+LMP$^*$ \cite{deng2018image} 
		& - & - & 57.7 & 26.7 & 46.4 & 26.2 & - & - 
		\\
		HHL$^*$ \cite{zhong2018generalizing}
		& - & - & 62.2 & 31.4 & 46.9 & 27.2 & - & - 
		\\
		\changed{DASy$^*$ \cite{bak2018domain}}
		& \changed{-} & \changed{-} & \changed{65.7} & \changed{-} 
		& \changed{-} & \changed{-} & \changed{-} & \changed{-}
		\\
		\hline
		\textbf{TAUDL} \cite{li2018Unsupervised}                                	
		& 44.7 & 31.2
		& 63.7 & 41.2
		& 61.7 & 43.5
		& 28.4	& 12.5
		\\
		\textbf{UTAL}                                	
		&\bf 56.3 		&\bf 42.3
		&\bf 69.2 		&\bf 46.2
		&\bf 62.3 		&\bf 44.6 
		&\bf 31.4		&\bf 13.1		\\
		\hline
		\changed{GCS \cite{chen2018group}({\em Supervised})}	
		& \changed{88.8}	& \changed{97.2}         
		& \changed{93.5}	& \changed{81.6}		
		& \changed{84.9} 	& \changed{69.5}
		& \changed{-} 		& \changed{-}
		\\ \hline
	\end{tabular}
	\vspace{-0.3cm}
\end{table*}

To further test realistic model performances, 
we introduced a new tracklet
person re-id benchmark based on DukeMTMC \cite{ristani2016performance}.
\changed{It differs from all the existing DukeMTMC variants
\cite{zheng2017unlabeled,wu2018exploit,gou2017dukemtmc4reid} 
by uniquely 
providing 
{\em automatically} generated tracklets.}
%
We built this new tracklet person re-id dataset as follows.
We first deployed an efficient deep learning tracker
that leverages a COCO+PASCAL trained {\em SSD} \cite{liu2016ssd} for pedestrian detection 
and an ImageNet trained {\em Inception}
\cite{szegedy2015going} for person appearance matching. 
Applying this tracker to all 
DukeMTMC raw videos, 
we generated 19,135 person tracklets.
Due to the inevitable detection and tracking errors caused by 
background clutters and visual ambiguity, 
these tracklets may present typical mistakes (e.g. ID switch) and corruptions (e.g. occlusion).
\changed{We name this test \textbf{\em DukeMTMC-SI-Tracklet},
abbreviated as \textbf{\em DukeTracklet}.} The DukeMTMC-SI-Tracklet dataset is publicly released at: {\em https://github.com/liminxian/DukeMTMC-SI-Tracklet}.

For benchmarking DukeTracklet, 
we need the ground-truth person ID labels of tracklets.
%
To this end, we used the criterion of spatio-temporal average Intersection over Union (IoU)
between detected tracklets and ground-truth trajectories available in DukeMTMC.
In particular,
we labelled an auto-generated tracklet by the 
ground-truth person ID associated with a manually-generated trajectory 
if their average IoU is over 50\%. 
Otherwise, we labelled the auto-generated tracklet as ``unknown ID''. 
To maximise the comparability with existing DukeMTMC variants,
we threw away those tracklets labelled with unknown IDs.
We finally 
obtained 12,647 person tracklets from 1,788 unique IDs.
The average tracklet duration is 65.9 frames.
To match DukeMTMC-ReID \cite{zheng2017unlabeled}, 
we set the same 702 training IDs
with the remaining 1,086 people for performance test
(missing 14 test IDs against DukeMTMC-ReID due to tracking failures).

\noindent \textbf{Tracklet Label Assignment. }
For each video re-id dataset, 
we simply assigned each tracklet with a unique label
in a camera-independent manner (Sec. \ref{sec:tracklet_labelling}).
For each multi-shot image datasets, 
we assumed all person images per ID per camera were drawn from 
a single pedestrian tracklet,
and similarly labelled them
as the video datasets.

\noindent \textbf{Performance Metrics. } 
We adopted the common Cumulative Matching Characteristic (CMC) and mean Average Precision (mAP) metrics \cite{zheng2015scalable} for model performance measurement. 

\noindent \textbf{Implementation Details. }
We used an ImageNet pre-trained ResNet-50 \cite{he2016deep} as the backbone net
for UTAL, along with 
an additional 2,048-D fully-connected (FC) layer 
for deriving the camera-shared representations.
Every camera-specific branch was formed by one FC classification layer.
Person bounding box images were resized to $256\!\times\!128$. 
To ensure each training mini-batch has person images 
from all cameras,
we set the batch size to 128 
for PRID2011, iLIDS-VID and CUHK03, 
and 384 
for MSMT17, Market-1501, MARS, DukeMTMC-ReID, and DukeTracklet.
In order to balance the model training speed across cameras,
we randomly selected the same number of tracklets
per camera and the same number of frame images (4 images) per
chosen tracklet when sampling each mini-batch.
We adopted the Adam optimiser \cite{kingma2014adam} with the learning rate of $3.5 \!\times\! 10^{-4}$
and the epoch of $200$.
By default, we set $\lambda\!=\!10$ for Eq~\eqref{eq:loss_UTAL},
$\alpha\!=\!1$ for Eq~\eqref{eq:tracklet_feat},
and $K\!=\!4$ for Eq~\eqref{eq:sigma}
in the following experiments.

\subsection{Comparisons to the State-Of-The-Art Methods}
\label{sec:exp_SOTA}

We compared two different sets of state-of-the-art methods
on image and video re-id datasets, due to the independent studies on them 
in the literature.

\begin{table*}[h]
	\centering
	\setlength{\tabcolsep}{0.07cm}
	\caption{Unsupervised person re-id on video based datasets.
		$^*$: Assume no tracking fragmentation.
		$^\dagger$: Use some ID labels for model initialisation.
	}
	\vskip -0.3cm
	\label{tab:vide_SOTA}
	\begin{tabular}
		{l||c|c|c||c|c|c||c|c|c|c||c|c|c|c}
		\hline
		Dataset		    						
		& \multicolumn{3}{c||}{PRID2011 \cite{hirzer2011person}}      	
		& \multicolumn{3}{c||}{iLIDS-VID \cite{wang2014person}}		
		& \multicolumn{4}{c||}{MARS \cite{zheng2016mars }}
		& \multicolumn{4}{c}{DukeTracklet }
		\\ \hline\hline
		Metric (\%)
		& Rank-1 	& Rank-5	& Rank-20					& Rank-1 	& Rank-5	& Rank-20
		& Rank-1 	& Rank-5	& Rank-20 	& mAP			& Rank-1 	& Rank-5	& Rank-20 	& mAP
		\\ \hline \hline
		GRDL \cite{kodirov2016person}		
		& 41.6	& 76.4	& 89.9					& 25.7	& 49.9	& 77.6
		& 19.3	& 33.2	& 46.5	& 9.6			& -	& -	& -	& -
		\\ 
		UnKISS \cite{khan2016unsupervised}				
		& 58.1	& 81.9	& 96.0					& 35.9	& 63.3	& 83.4
		& 22.3	& 37.4	& 53.6	& 10.6			& -	& -	& -	& -
		\\ \hline
		SMP$^*$ \cite{liu2017stepwise}
		& 80.9	& 95.6	& 99.4					& 41.7	& 66.3	& 80.7
		& 23.9   	& 35.8	& 44.9	& 10.5 			& -	& -	& -	& - 
		\\ 
		DGM+MLAPG$^\dagger$ \cite{ye2017dynamic}     
		& 73.1	& 92.5	& 99.0 					& 37.1	& 61.3	& 82.0
		& 24.6   	& 42.6	& 57.2	& 11.8 			& -	& -	& -	& - 
		\\ 
		DGM+IDE$^\dagger$ \cite{ye2017dynamic}
		& 56.4	& 81.3	& 96.4 					& 36.2	& 62.8	& 82.7
		& 36.8  & 54.0  & 68.5 & 21.3				& -	& -	& -	& - 
		\\
		RACE$^\dagger$ \cite{ye2018robust}
		& 50.6	& 79.4	& 91.8 					& 19.3	& 39.3	& 68.7
		& 43.2  & 57.1  & 67.6 & 24.5				& -	& -	& -	& - 
		\\
		\changed{DASy$^*$ \cite{bak2018domain}}
		& \changed{43.0} & \changed{-} & \changed{-} & \changed{56.5} & \changed{-} & \changed{-}
		& \changed{-} & \changed{-} & \changed{-} & \changed{-} 	
		& \changed{-} & \changed{-} & \changed{-} & \changed{-}
		\\ 
		\hline
		DAL \cite{chen2018deep}
		& \bf 85.3	& \bf 97.0	& \bf 99.6		& \bf 56.9	& \bf 80.6	& \bf 91.9
		& 46.8  & 63.9  & 77.5 & 21.4				& -	& -	& -	& - 
		\\ \hline
		\textbf{TAUDL} \cite{li2018Unsupervised} 
		& 49.4	& 78.7	& 98.9					& 26.7	& 51.3	& 82.0
		& 43.8 & 59.9 & 72.8 & 29.1 
		& 26.1 & 42.0 & 57.2	& 20.8 
		\\
		\textbf{UTAL}
		& 54.7	& 83.1	& 96.2
		& 35.1	& 59.0	& 83.8
		& \bf 49.9	& \bf 66.4 	& \bf 77.8 	& \bf 35.2
		& \bf 43.8 	& \bf 62.8	& \bf 76.5	& \bf 36.6
		\\ \hline
		\changed{Snippet \cite{chen2018video}({\em Supervised})}
		& \changed{93.0} & \changed{99.3} & \changed{100.0}
		& \changed{85.4} & \changed{96.7} & \changed{99.5}
		& \changed{86.3} & \changed{94.7} & \changed{98.2} & \changed{76.1}
		& \changed{-} & \changed{-} & \changed{-} & \changed{-}
		\\ \hline
	\end{tabular}
	\vspace{-0.3cm}
\end{table*}

\noindent {\bf Evaluation on Image Datasets. }
Table \ref{tab:img_SOTA}
shows the unsupervised re-id performance
of the proposed UTAL and \changed{15} state-of-the-art methods including 
3 hand-crafted feature based methods 
(Dic \cite{kodirov2015dictionary}, ISR \cite{lisanti2015person}, RKSL \cite{wang2016towards}) and 
\changed{12} auxiliary knowledge (identity/attribute) transfer based models 
(AE \cite{lee2008sparse},
AML \cite{ye2007adaptive}, 
UsNCA \cite{qin2015unsupervised}, 
CAMEL \cite{yu2017cross},
JSTL \cite{xiao2016learning},
PUL\cite{fan2017unsupervised},
TJ-AIDL \cite{want2018Transfer},
CycleGAN \cite{zhu2017unpaired},
SPGAN \cite{deng2018image},
HHL \cite{zhong2018generalizing},
\changed{DASy \cite{bak2018domain}}).
%
The results show four observations as follows.

\noindent{\bf(1)} 
Among the existing methods,
the knowledge transfer based models are often superior
due to the use of {\em additional} label information,
e.g. Rank-1 39.4\% by CAMEL {\em vs} 36.5\% by Dic on CUHK03; 
\changed{65.7\% by DASy} {\em vs} 50.2\% by Dic on
Market-1501.
To that end, CAMEL needs to benefit from learning on $7$ different
person re-id datasets of diverse domains
(CUHK03 \cite{li2014deepreid}, CUHK01 \cite{li2012human}, PRID \cite{hirzer2011person}, VIPeR \cite{gray2008viewpoint}, 
i-LIDS \cite{prosser2010person})
including a total of 44,685 images and 3,791 IDs;
HHL requires to utilise labelled Market-1501 (750 IDs) or DukeMTMC-ReID (702 IDs) as 
the source training data.
\changed{DASy needs elaborative ID synthesis
and adaptation.} 

\noindent
{\bf (2)} The proposed UTAL outperforms all competitors with significant margins.
For example, the Rank-1 margin by UTAL over HHL
is 7.0\% (69.2-62.2) on Market-1501 and 15.4\% (62.3-46.9) on DukeMTMC-ReID.
Also, our preliminary method TAUDL already surpasses all previous methods.
It is worth pointing out that UTAL dose not benefit from any
additional labelled source domain training data as compared to the strong
alternative HHL.
Importantly, UTAL is potentially more scalable due to {\em no} reliance at all 
on the similarity constraint between source and target domains.

\noindent{\bf (3)} 
The UTAL is simpler to train
with a simple end-to-end model learning, 
{\em vs} the alternated
deep CNN training and data clustering required by PUL, 
a two-stage model training of TJ-AIDL, 
high GAN training difficulty of HHL,
\changed{and elaborative ID synthesis of DASy.}
%
These results show both the performance advantage and 
model design superiority of UTAL 
over 
state-of-the-art re-id methods.

\noindent{\bf (4)} 
\changed{A large performance gap exists between
unsupervised and supervised learning models. Further improvement is
required on unsupervised learning algorithms.}

\noindent {\bf Evaluation on Video Datasets. }
In Table \ref{tab:vide_SOTA}, we compared the UTAL with \changed{8} 
state-of-the-art unsupervised video re-id models
(GRDL \cite{kodirov2016person},
UnKISS \cite{khan2016unsupervised},
SMP \cite{liu2017stepwise},
DGM+MLAPG/IDE \cite{ye2017dynamic},
DAL \cite{chen2018deep},
RACE \cite{ye2018robust},
\changed{DASy \cite{bak2018domain}})
on the video benchmarks.
Unlike UTAL, all these existing models except DAL 
are {\em not} end-to-end deep learning methods
using hand-crafted or independently trained deep features as input.

The comparisons show that, our UTAL outperforms all existing video person re-id models on the large scale video dataset MARS,
e.g. by a Rank-1 margin of 3.1\% (49.9-46.8) 
and a mAP margin of 13.8\% (35.2-21.4)
over the best competitor DAL. 
However, UTAL is
inferior than top existing models on the two small benchmarks
iLIDS-VID (300 training tracklets) and PRID2011 (178 training tracklets),
{\em vs} 8,298 training tracklets in MARS.
%
This shows that UTAL does need sufficient tracklet data 
in order to have its performance advantage. As the required
tracklet data are not manually labelled, this requirement is not a
hindrance to its scalability on large scale deployments. 
Quite the contrary,
UTAL works the best when large unlabelled video data are available.
A model would benefit 
from pre-training using UTAL on large auxiliary unlabelled videos
with similar viewing conditions.

\subsection{Component Analysis and Discussion}
\label{sec:exp_component}

We conducted 
detailed UTAL model component analysis
on two large tracklet re-id datasets,
MARS 
and DukeTracklet.

\noindent\textbf{Per-Camera Tracklet Discrimination Learning.} 
We started by testing the 
performance impact of the PCTD component.
This is achieved by designing a baseline that treats all cameras together, 
that is, concatenating the per-camera tracklet label spaces
and deploying the Cross-Entropy loss
for learning a \textbf{\em Single-Task Classification} (STC).
In this analysis, we did not consider the cross-camera tracklet association component
for a more focused evaluation.

Table \ref{tab:PCJL} shows that, 
the proposed PCTD design is significantly superior
over the STC learning algorithm,
e.g. achieving Rank-1 gain of 
27.9\% (43.8-15.9), and 27.8\% (31.7-3.9) on
MARS and DukeTracklet, respectively.
The results demonstrate modelling advantages of PCTD 
in exploiting unsupervised tracklet labels for
learning cross-view re-id discriminative features.
This validates the proposed idea of
implicitly deriving a cross-camera shared feature representation
through a multi-camera multi-task learning strategy.

\begin{table}[h]
	\centering
	\setlength{\tabcolsep}{0.39cm}
	\caption{Effect of Per-Camera Tracklet Discrimination  (PCTD) learning.
}
	\label{tab:PCJL}
	\vskip -0.3cm
	\begin{tabular}
		{c||c|c||c|c}
		\hline
		Dataset				
		& \multicolumn{2}{c||}{MARS \cite{zheng2016mars}}	
		& \multicolumn{2}{c}{DukeTracklet}		\\ \hline 
		Metric (\%)			& Rank-1	& mAP		& Rank-1	& mAP	
		\\ \hline \hline
		STC		& 15.9	& 10.0		& 3.9	& 4.7
		\\ \hline
		\bf PCTD	& \textbf{43.8}	& \textbf{31.4}	& \textbf{31.7}	& \textbf{26.4}
		\\ \hline
	\end{tabular}
\end{table}

Recall that we propose 
a soft label (Eq \eqref{eq:soft_label}) based Cross-Entropy loss
(Eq \eqref{eq:soft_CE_loss})
for tackling the notorious trajectory fragmentation challenge.
To test how much benefit our soft labelling strategy brings to
unsupervised tracklet re-id,
we compared it against the one-hot class {\em hard}-labelling counterpart (Eq \eqref{eq:CE_loss}).  
Table \ref{tab:soft_hard_label} shows that 
the proposed soft-labelling is significantly superior, 
suggesting a clear benefit in mitigating the negative impact of
trajectory fragmentation.
This is due to the intrinsic capability of 
exploiting the appearance pairwise affinity knowledge among tracklets per camera.

\begin{table}[h]
	\centering
	\setlength{\tabcolsep}{0.35cm}
	\caption{Soft-labelling {\em versus} hard-labelling.
	}
	\label{tab:soft_hard_label}
	\vskip -0.3cm
	\begin{tabular}
		{c||c|c||c|c}
		\hline
		Dataset		
		& \multicolumn{2}{c||}{MARS \cite{zheng2016mars}}	
		& \multicolumn{2}{c}{DukeTracklet}		\\ \hline 
		Metric (\%)			& Rank-1	& mAP		& Rank-1	& mAP
		\\ \hline \hline
		Hard-Labelling			& 35.5		& 20.5		& 24.5		& 17.3
		\\ \hline
		\bf Soft-Labelling		& \textbf{49.9}	& \textbf{35.2}	& \textbf{43.8}	& \textbf{36.6}
		\\ \hline
	\end{tabular}
\end{table}

\begin{figure} [h]
	\centering 
	\includegraphics[width=0.5\linewidth, height=2.4cm]
	{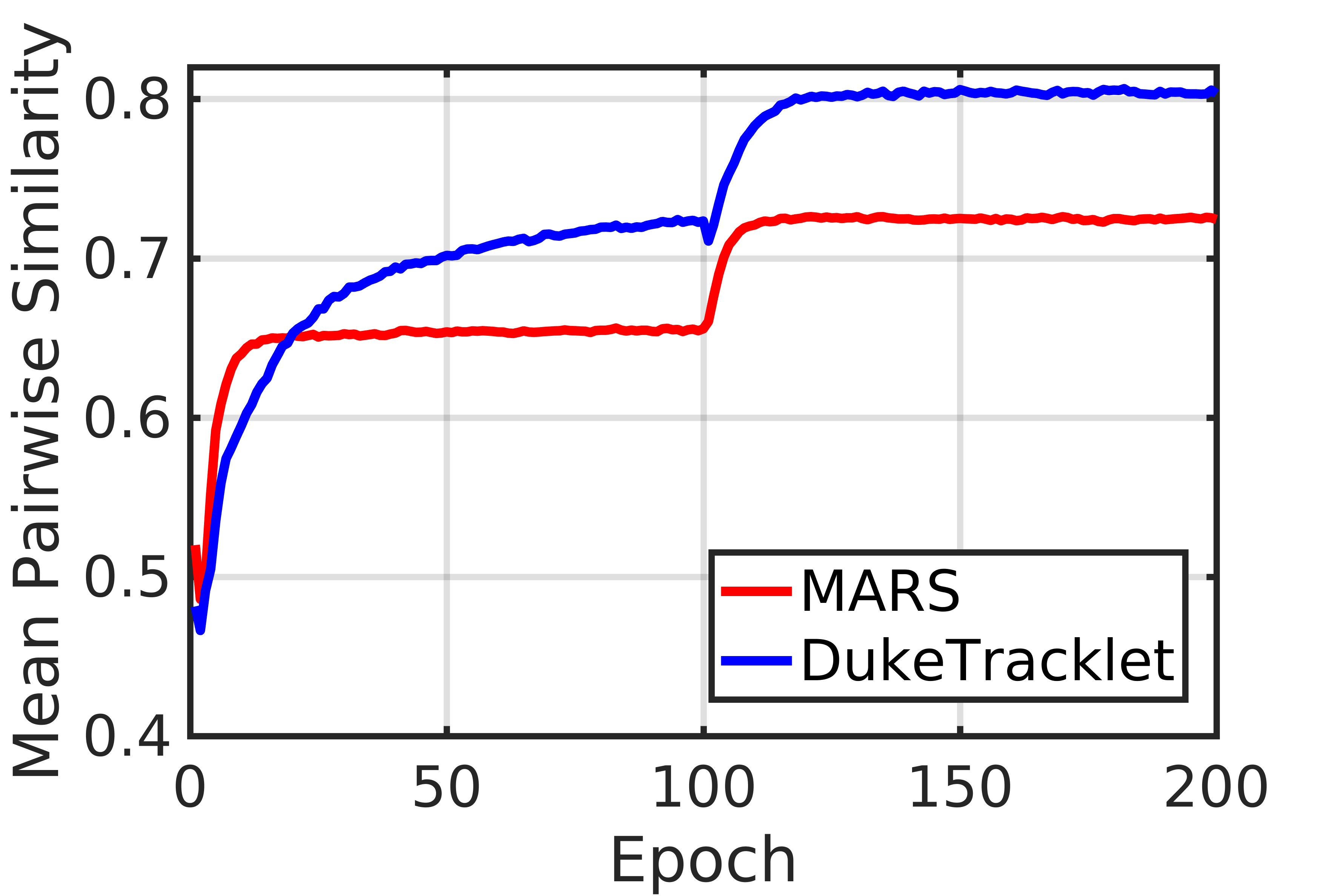}
	\vskip -0.3cm
	\caption{
		The evolving process of tracklet soft label quality 
		over the model training epochs on MARS and DukeTracklet.
	}
	\label{fig:soft_label_quality}
	\vspace{-0.3cm}
\end{figure}

We further examined the ID discrimination capability of tracklet soft labels
that underpins its outperforming over the corresponding hard labels.
To this end, 
we measured the {\em Mean Pairwise Similarity} (MPS) of soft label vectors 
assigned to per-camera same-ID tracklets. 
Figure \ref{fig:soft_label_quality} shows that,
the MPS metric goes higher as the training epoch increases,
particularly after the CCTA loss is further exploited 
in the middle of training (at 100$^\text{th}$ epoch).
This indicates explicitly the evolving process of
mining the discriminative knowledge among fragmented tracklets
with the same person ID labels in a self-supervising fashion.

In effect, the affinity measurement (Eq \eqref{eq:sparse}) 
used for computing soft labels 
can be 
useful for automatically merging
short fragmented tracklets into long trajectories per camera.
We tested this {\em tracking refinement} capability of our method. 
In particular,
we built a sparse connection graph by thresholding 
the pairwise affinity scores at $0.5$, 
analysed the connected tracklet components
\cite{pearce2005improved},
and merged all tracklets in each component 
into a long trajectory.
%
%

\begin{figure} [h]
	\centering 
	\includegraphics[width=\linewidth]{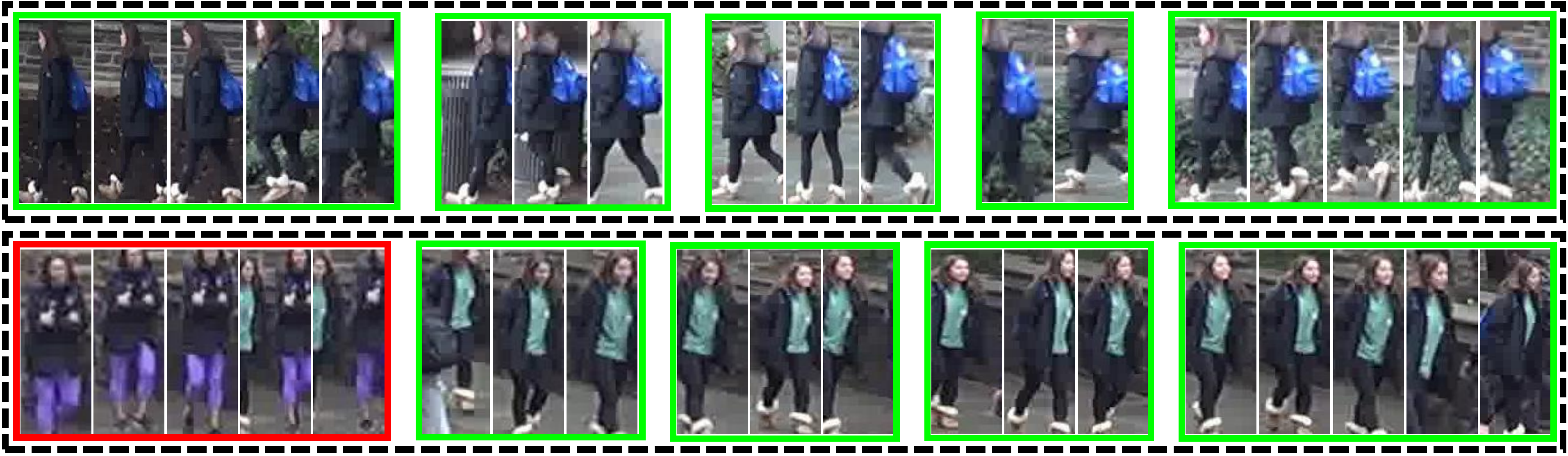}
	\vskip -0.3cm
	\caption{
		Example long trajectories discovered 
		by UTAL among unlabelled short fragmented tracklets.
		Each row denotes a case.
		%
		The tracklets in green/red bounding box denote
		the true/false matches, respectively.
		Failure tracklet merging may be due to detection and tracking 
		errors.
	}
	\label{fig:long_track}
	\vspace{-0.2cm}
\end{figure}

Table \ref{tab:tracking_ref} shows that 
with our per-camera tracklet affinity measurement,
even such a simple strategy can merge 
4,389/2,527 out of 8,298/5,803 short tracklets 
into 1,532/982 long trajectories
at the NMI (Normalised Mutual Information)
rate of 0.896/0.934
on MARS/DukeTracklet.
This not only suggests the usefulness of UTAL
in tracklet refinement,
but also reflects the underlying correlation
between tracking and person re-id. 
%
For visual quality examination, 
we gave example cases for tracking refinement
in Fig. \ref{fig:long_track}. 

\begin{table}[h]
	\centering
	\setlength{\tabcolsep}{0.5cm}
	\caption{Evaluating the tracking refinement capability of 
		soft labels.
	}
	\label{tab:tracking_ref}
	\vskip -0.3cm
	\begin{tabular}
		{c||c||c}
		\hline
		Dataset		
		& \multicolumn{1}{c||}{MARS \cite{zheng2016mars}}	
		& \multicolumn{1}{c}{DukeTracklet}		
		\\ \hline \hline
		Original Tracklets 	& 8,298 	& 5,803
		\\ \hline\hline
		Mergable Tracklets 			& 4,389  	& 2,527
		\\ \hline
		Long Trajectories 			& 1,532 	& 928	
		\\ \hline \hline
		NMI							& 0.896		& 0.934
		\\ \hline
	\end{tabular}
	\vspace{-0.3cm}
\end{table}

Algorithmically, our soft label PCTD 
naturally inherits the cross-class (ID) knowledge transfer capability 
from {\em Knowledge Distillation} \cite{hinton2015distilling}.
It is interesting to see how much performance benefit this can bring 
to unsupervised tracklet re-id. 
%
To this end, we conducted a controlled experiment 
with only one randomly selected training tracklet per ID per camera.
Doing so enforces that no multiple per-camera tracklets
share the same person ID,
%
which more explicitly evaluates the impact of cross-ID knowledge transfer.  
Note, this leads to probably inferior re-id model generalisation
capability due to less
training data used in optimisation. 
Table \ref{tab:extra_effect_soft_label} shows that 
the cross-ID knowledge transfer gives notable performance improvements. 

\begin{table}[h]
	\centering
	\setlength{\tabcolsep}{0.4cm}
	\caption{Evaluating the cross-ID knowledge transfer effect 
		of soft labels.
	}
	\label{tab:extra_effect_soft_label}
	\vskip -0.3cm
	\begin{tabular}
		{c||c|c||c|c}
		\hline
		Dataset		
		& \multicolumn{2}{c||}{MARS \cite{zheng2016mars}}	
		& \multicolumn{2}{c}{DukeTracklet}		\\ \hline 
		Metric (\%)			& Rank-1	& mAP		& Rank-1	& mAP
		\\ \hline \hline
		Hard Label			& 45.1		& 31.1		& 28.6		& 20.8
		\\ \hline
		\bf Soft Label		& \textbf{46.5}	& \textbf{31.2}	& \textbf{31.7}	& \textbf{24.8}
		\\ \hline
	\end{tabular}
\end{table}

\noindent \textbf{Cross-Camera Tracklet Association Learning.} 
We evaluated the CCTA component by 
measuring the performance drop once eliminating it.
Table \ref{tab:TA} shows that CCTA brings a significant re-id accuracy benefit, 
e.g. a Rank-1 boost of 
6.1\% (49.9-43.8) and 12.1\% (43.8-31.7)
on MARS and DukeTracklet, respectively.
This suggests the importance of cross-camera ID class correlation modelling 
and the capability of our CCTA formulation
in reliably associating tracklets across cameras
for unsupervised re-id model learning.

\begin{table}[h]
	\centering
	\setlength{\tabcolsep}{0.43cm}
	\caption{Effect of Cross-Camera Tracklet Association (CCTA) learning.}
	\label{tab:TA}
	\vskip -0.3cm
	\begin{tabular}
		{c||c|c||c|c}
		\hline
		Dataset				
		& \multicolumn{2}{c||}{MARS \cite{zheng2016mars}}
		& \multicolumn{2}{c}{DukeTracklet}
		\\ \hline
		CCTA		& Rank-1		& mAP			& Rank-1		& mAP	
		\\ \hline \hline
		\xmark		& 43.8			& 31.4			& 31.7			& 26.4
		\\ \hline
		\cmark		& \bf{49.9}  		& \bf{35.2}		& \bf{43.8}		& \bf{36.6}
		\\ \hline
	\end{tabular}
\end{table}

\begin{figure} [h]
	\centering
	\subfigure[Number of tracklet pairs.]
	{
		\includegraphics[width=0.45\linewidth, , height=2.2cm]{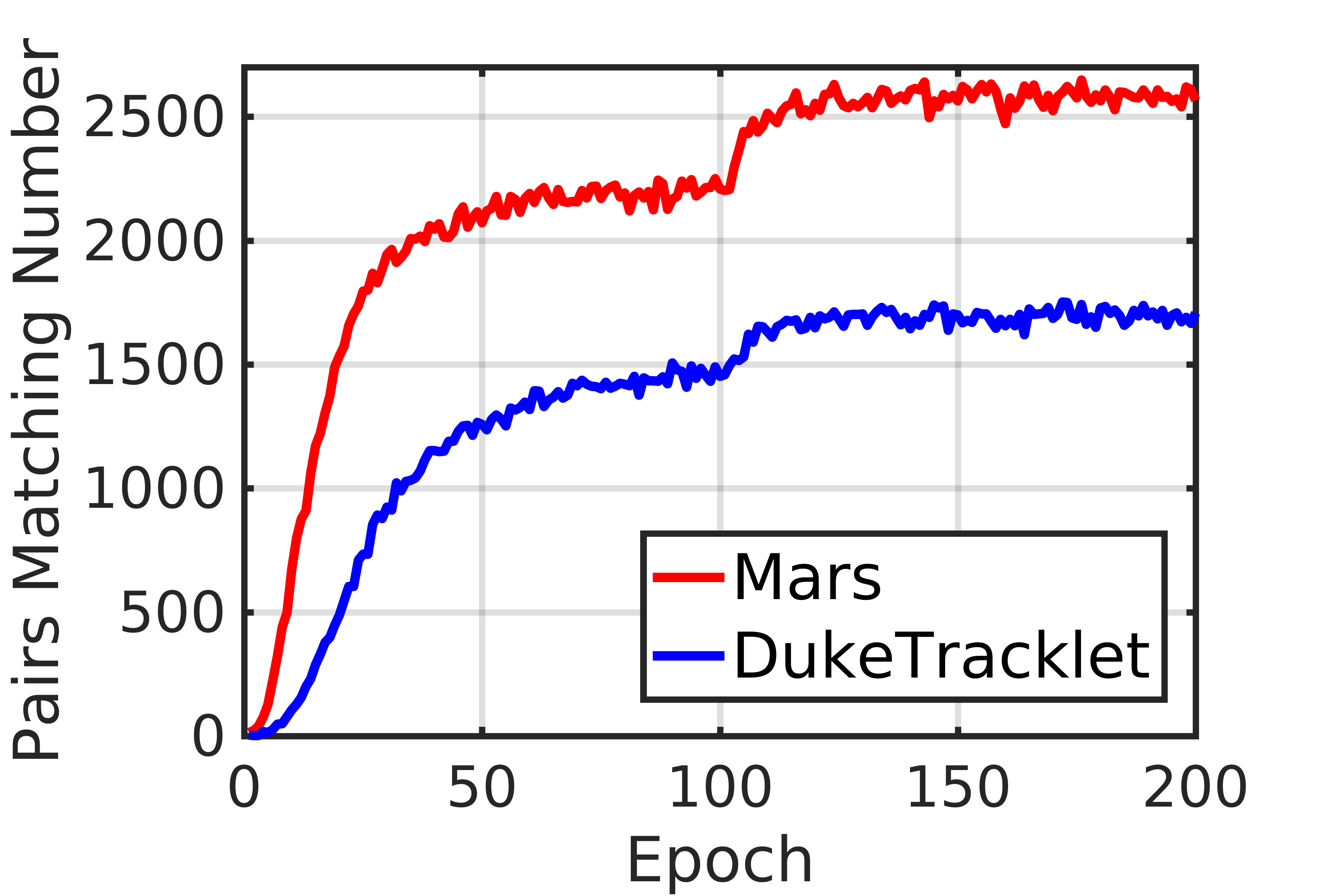}
	}
	\subfigure[Precision of matching pairs.]
	{
		\includegraphics[width=0.45\linewidth, , height=2.2cm]{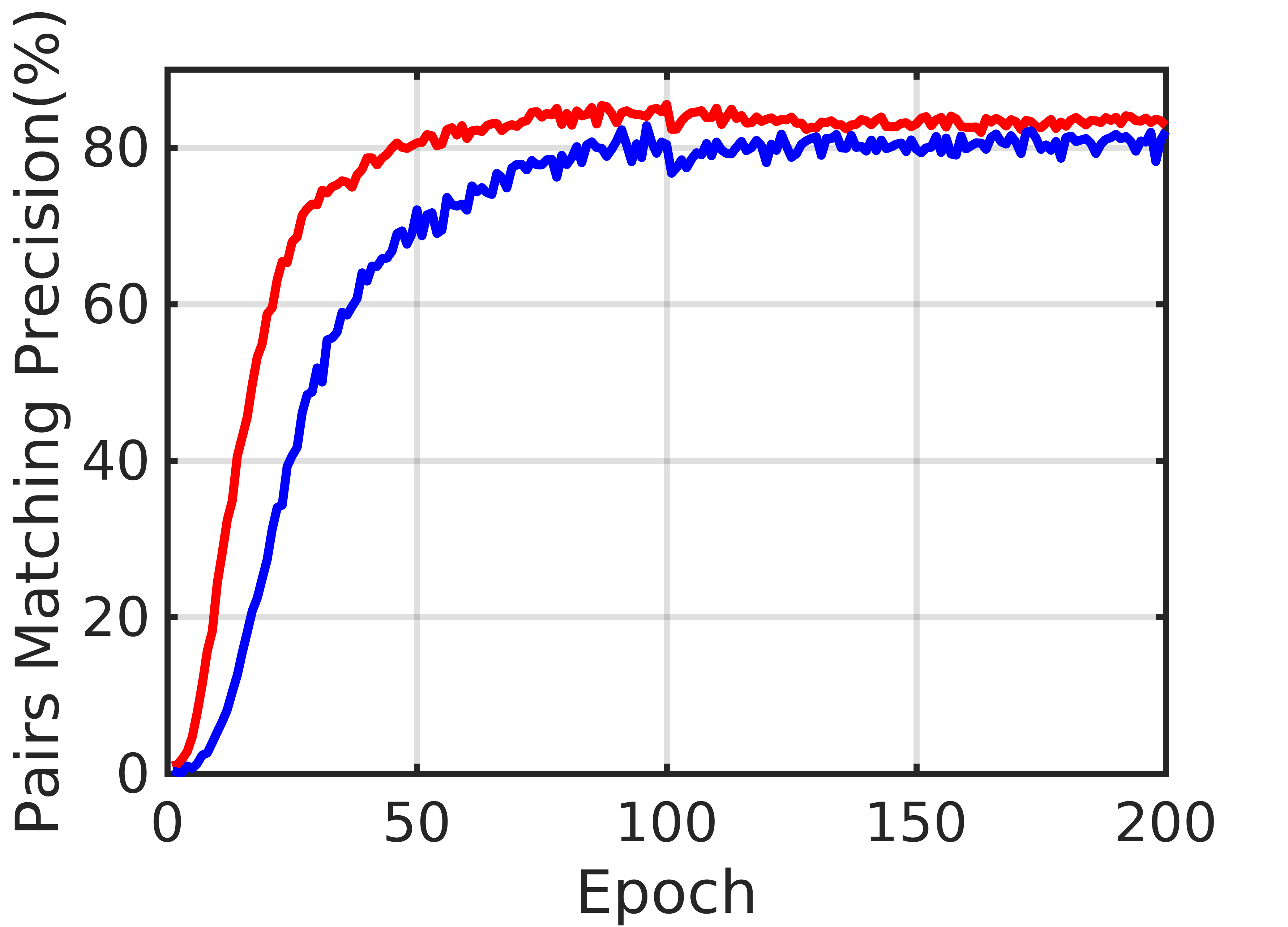}
	}
	\vskip -0.3cm
	\caption{
		The evolving process of self-discovered cross-camera tracklet matching pairs
		in {\bf (a)} number and {\bf (b)} precision  
		throughout the training.
	}
	\label{fig:CCTA_track_pairs}
\end{figure}

To further examine why CCTA enables 
more discriminative re-id model learning,
we tracked the self-discovered cross-camera tracklet matching pairs
throughout the training. 
Figure \ref{fig:CCTA_track_pairs} shows that both the number and precision
of self-discovered cross-camera tracklet pairs 
increase. 
This echoes the model performance superiority of UTAL.

\begin{figure} [h]
	\centering 
	\subfigure[The sensitive of  $\alpha$]{
		\includegraphics[width=0.3\linewidth]{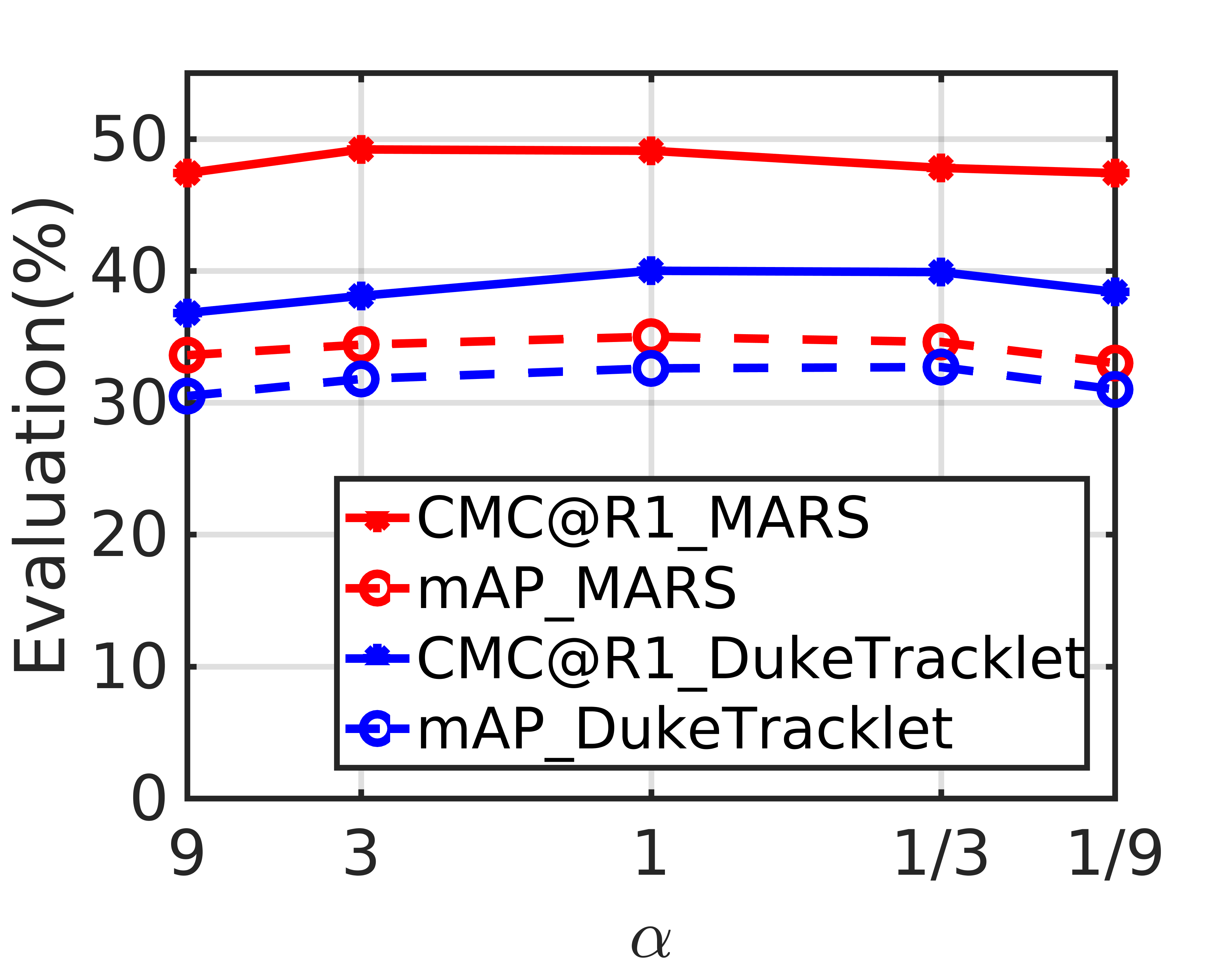}
	}
	\subfigure[The sensitive of $K$]{
		\includegraphics[width=0.3\linewidth]{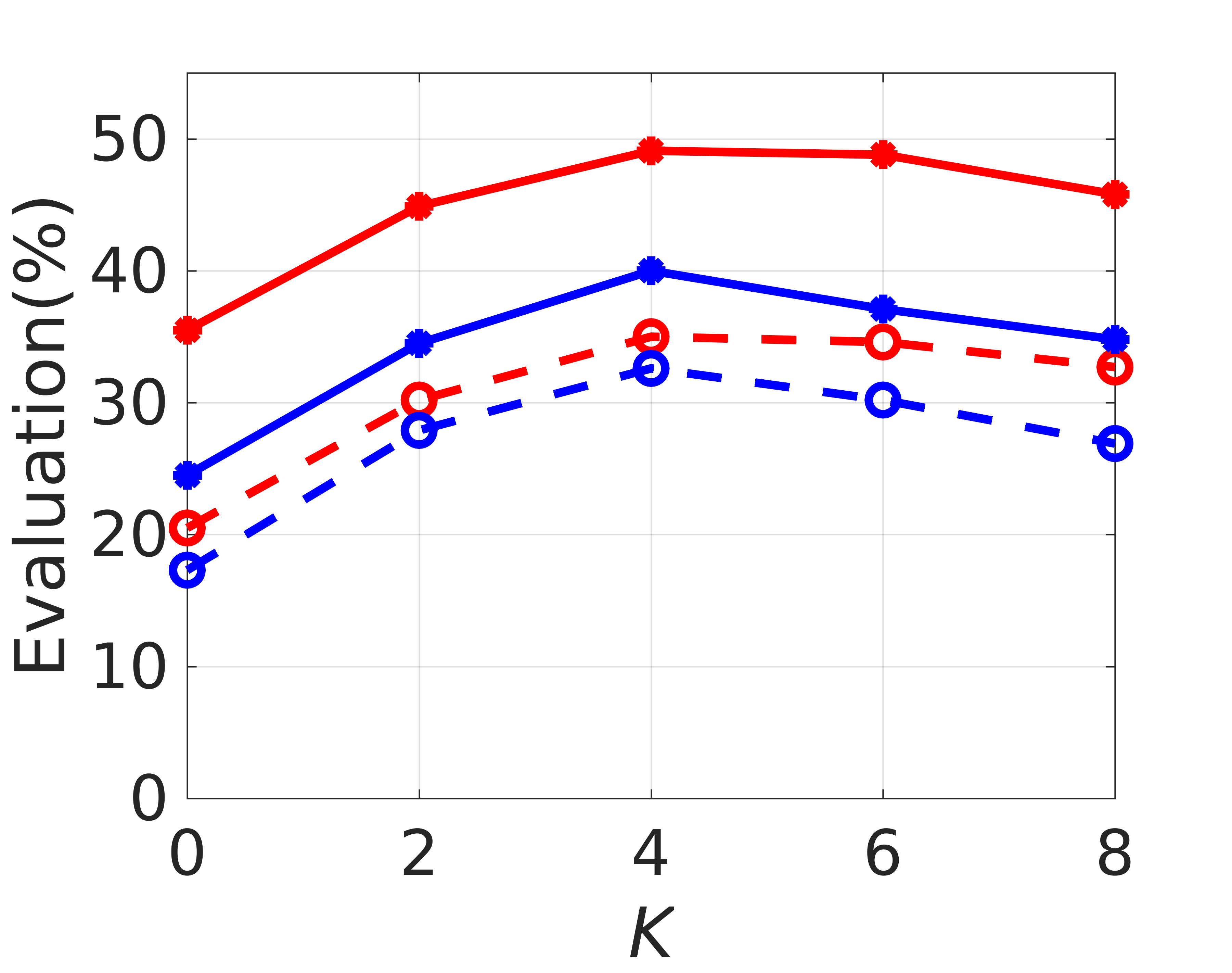}
	}
	\subfigure[The sensitive of  $\lambda$]{
		\includegraphics[width=0.3\linewidth]{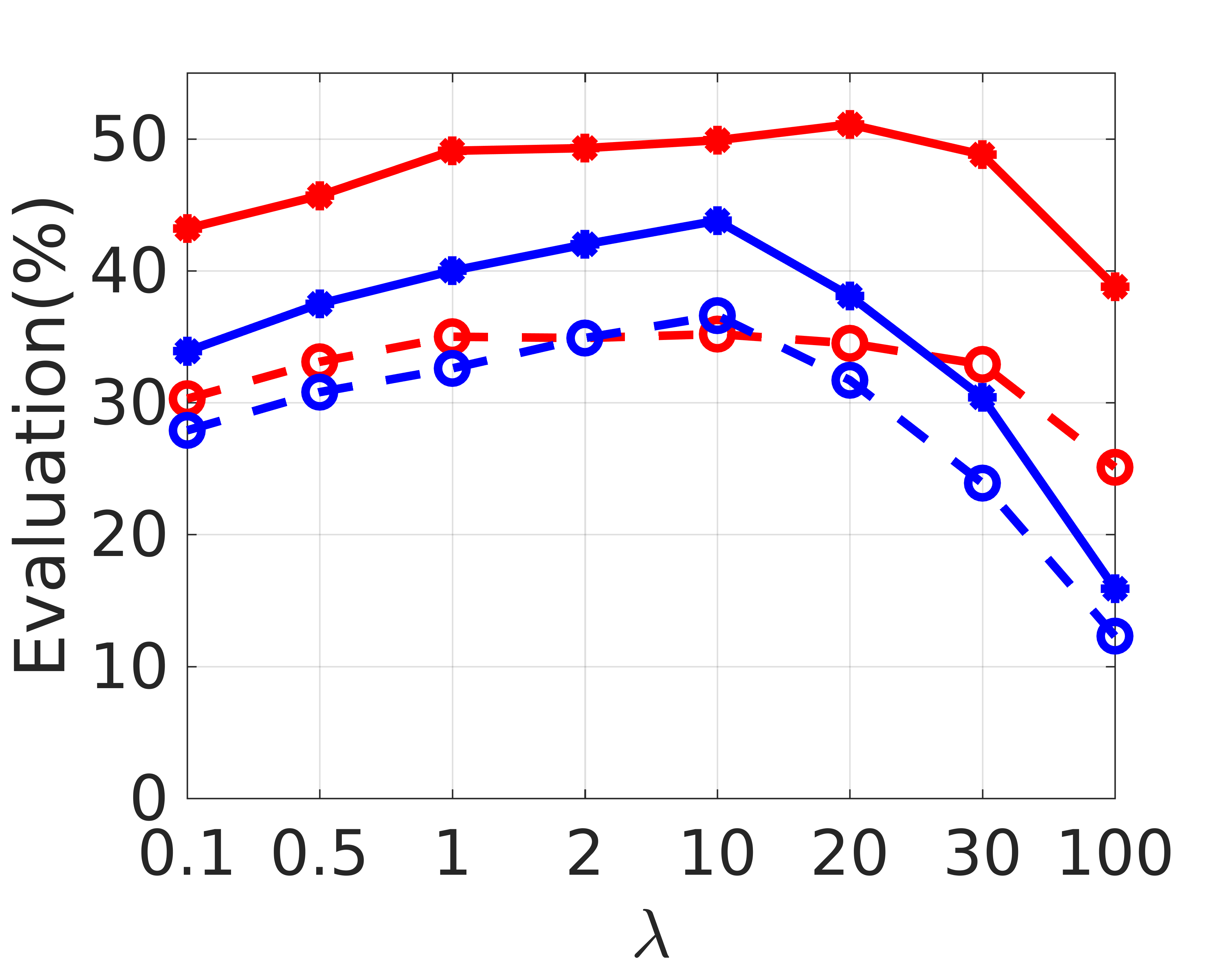}
	}
	\vskip -0.3cm
	\caption{
		Analysis of the UTAL model parameters.
	}
	\vspace{-0.2cm}
	\label{fig:para_analysis}
\end{figure}

\noindent \textbf{Model Parameters.} 
We evaluated the performance impact 
of three UTAL hyper-parameters:
(1) the tracklet feature update learning rate $\alpha$ (Eq \eqref{eq:tracklet_feat}),
(2) the sparsity $K$ of the tracklet affinity matrix used in computing the soft labels (Eq \eqref{eq:sparse} and \eqref{eq:sigma});
(3) the loss balance weight $\lambda$ (Eq \eqref{eq:loss_UTAL}).
Figure \ref{fig:para_analysis} shows that:
(1) 
$\alpha$ is not sensitive with a wide satisfactory range.
This suggests a stable model learning procedure.
(2) 
 $K$ has an optimal value at ``4''.
Too small values lose the opportunities of incorporating same-ID tracklets into the soft labels whilst the opposite instead introduces distracting/noisy neighbour information.
(3) 
$\lambda$ is found more domain dependent,
with the preference values around ``10''.
This indicates a higher importance of 
cross-camera tracklet association and matching.

\noindent \textbf{Cross-Camera Nearest Neighbours (CCNN).} 
\changed{
We evaluated the effect of CCNN $\mathcal{R}$ 
(Eq \eqref{eq:rep_knn}) used in the CCTA loss.
We compared two designs: Our reciprocal 2-way $1$-NN {\em vs.} common 1-way $1$-NN.
Table \ref{tab:NN_analysis} shows that 
the more strict 2-way $1$-NN gives better overall performance
whilst the 1-way $1$-NN has a slight advantage in Rank-1 on
MARS (50.5\% vs. 49.9\%).
By 2-way, we found using more neighbours (5/10-NN) degrades
model performance. 
This is due to the introduction of more false cross-camera matches.}

\begin{table}[h]
	\centering
	\setlength{\tabcolsep}{0.38cm}
	\caption{Effect of cross-camera nearest neighbours.}
	\label{tab:NN_analysis}
	\vskip -0.3cm
	\begin{tabular}
		{c||c|c||c|c}
		\hline
		Dataset
		& \multicolumn{2}{c||}{MARS \cite{zheng2016mars}}	
		& \multicolumn{2}{c}{DukeTracklet}		\\ \hline 
		Metric (\%)			
		& Rank-1	& mAP
		& Rank-1	& mAP
		\\ \hline \hline
		1-way $1$-NN				
		&\bf 50.5	&33.2 		
		&41.9	&34.5
		\\ \hline \hline
		2-way $1$-NN		
		&49.9	&\bf 35.2		
		&\bf 43.8	&\bf 36.6
		\\ \hline
		2-way	$5$-NN	
		&47.6	& 34.5		 
		& 38.0	& 31.6
		\\ \hline
		2-way	$10$-NN	
		&45.5 &32.5 &33.9 &27.1
		\\ \hline
	\end{tabular}
	\vspace{-0.2cm}
\end{table}

\noindent \textbf{Tracklet Sampling {\em versus} All Tracklets. }
In our preliminary solution TAUDL \cite{li2018Unsupervised}, 
we considered
a Sparse Space-Time Tracklet (SSTT) sampling strategy
instead of unsupervised learning on all tracklet data.
It is useful in minimising the person ID duplication rate in tracklets.
However, 
such a data sampling throws away 
a large number of tracklets with rich information of person appearance
exhibited continuously and dynamically over space and time.
To examine this, we compared the SSTT sampling
with using all tracklets. 

Table \ref{tab:sampling_All} shows two observations:
(1) In overall re-id performance, our preliminary method {TAUDL} \cite{li2018Unsupervised} 
is outperformed significantly by UTAL. 
For example, 
the Rank-1 results are improved by 6.1\% (49.9-43.8) 
on MARS and by 17.7\% (43.8-26.1) on DukeTracklet.
(2) When using the same UTAL model, 
the SSTT strategy leads to inferior re-id rates as compared with using all tracklets.
For instance, the Rank-1 performance drop is 4.8\% (49.9-45.1) on MARS,
and 12.1\% (43.8-31.7) on DukeTracklet.
%
These performance gains are 
due to the proposed soft label learning idea that
effectively handles the trajectory fragmentation problem.
%
Overall, this validates the efficacy of our model design
in solving the SSTT's limitation 
whilst more effectively tackling the ID duplication 
(due to trajectory fragmentation) problem.

\begin{table}[h]
	\centering
	\setlength{\tabcolsep}{0.27cm}
	\caption{Tracklet sampling {\em versus} using all tracklets.
	}
	\vskip -0.3cm
	\label{tab:sampling_All}
	\begin{tabular}
		{c||c|c||c|c}
		\hline
		Dataset			
		& \multicolumn{2}{c||}{MARS \cite{zheng2016mars}}
		& \multicolumn{2}{c}{DukeTracklet }
		\\ \hline\hline
		Metric (\%)
		& Rank-1	& mAP			& Rank-1	& mAP
		\\ \hline \hline
		{TAUDL} \cite{li2018Unsupervised} 
		& 43.8	& 29.1
		& 26.1	& 20.8 
		\\ \hline
		\textbf{UTAL(SSTT)}
		& 45.1	& 31.1
		& 31.7	& 24.8
		\\
		\textbf{UTAL(All Tracklets)}
		& \textbf{49.9}	& \textbf{35.2} 
		& \textbf{43.8}	& \textbf{36.6} 
		\\ \hline
	\end{tabular}
	\vspace{-0.2cm}
\end{table}

\noindent \textbf{Effect of Neural Network Architecture. }
The model generalisation performance of UTAL
may depend on the selection of neural network architecture.
To assess this aspect, we evaluated
one more UTAL variant using 
a more recent DenseNet-121 \cite{huang2017densely}
as the backbone network,
{\em versus} the default choice ResNet-50 \cite{he2016deep}.
Table \ref{tab:backbone} shows that 
even superior re-id performances can be obtained
when using a stronger network architecture. 
This suggests that UTAL can readily benefit from the advancement of network designs.

\begin{table}[h]
	\centering
	\setlength{\tabcolsep}{0.3cm}
	\caption{Effect of backbone neural network in UTAL.
	}
	\vskip -0.3cm
	\label{tab:backbone}
	\begin{tabular}
		{c||c|c||c|c}
		\hline
		Dataset			
		& \multicolumn{2}{c||}{MARS \cite{zheng2016mars}}
		& \multicolumn{2}{c}{DukeTracklet }
		\\ \hline\hline
		Metric (\%)
		& Rank-1	& mAP			& Rank-1	& mAP
		\\ \hline \hline
		ResNet-50 \cite{he2016deep}
		& 49.9	& 35.2
		& 43.8	& 36.6
		\\ \hline
		DenseNet-121 \cite{huang2017densely}
		& \textbf{51.6}		& \textbf{35.9}
		& \textbf{44.3}		& \textbf{36.7}
		\\ \hline
	\end{tabular}
	\vspace{-0.2cm}
\end{table}

\noindent \textbf{Weakly Supervised Tracklet Association Learning. }
For training data labelling in person re-id,
the most costly procedure is on exhaustive manual search
of cross-camera image/tracklet matching pairs.
It is often unknown where and when 
a specific person will appear given complex 
camera spatio-temporal topology 
and unconstrained people's behaviours
in the public spaces. 
Therefore, per-camera independent ID labelling is 
more affordable. 
Such labelled data
are much weaker and less informative, 
due to the lack of cross-camera positive and negative ID pairs information.
We call the setting 
{\em Weakly Supervised Learning} (WSL).

The proposed UTAL model can be flexibly applied
in the WSL setting.
Interestingly, this allows to test how much re-id performance benefit
such labels can provide.
Unlike in the unsupervised learning setting,
the soft label based PCTD loss is {\em no longer} necessary
in WSL given the within-camera ID information.
Hence, we instead deployed the hard one-hot label (Eq \eqref{eq:CE_loss}) based PCTD loss (Eq \eqref{eq:PCTD_loss}).
Table \ref{tab:WSL} shows that 
such weak labels are informative and useful for person re-id
by the UTAL method.
This test indicates 
a wide suitability and usability of our method
in practical deployments under various labelling budgets. 

\begin{table}[h]
	\centering
	\setlength{\tabcolsep}{0.27cm}
	\caption{Evaluation of weakly supervised tracklet association learning.
	}
	\label{tab:WSL}
	\vskip -0.3cm
	\begin{tabular}
		{c||c|c||c|c}
		\hline
		Dataset			
		& \multicolumn{2}{c||}{MARS \cite{zheng2016mars}}
		& \multicolumn{2}{c}{DukeTracklet}
		\\ \hline
		Metric (\%)			& Rank-1	& mAP			& Rank-1	& mAP	
		\\ \hline \hline
		Unsupervised		& 49.9		& 35.2			& 43.8		& 36.6
		\\ \hline
		Weakly Supervised	& \textbf{59.5}  & \textbf{51.7}	& \textbf{46.4}	& \textbf{39.0}
		\\ \hline
	\end{tabular}
	\vspace{-0.1cm}
\end{table}

\noindent {\bf Manual Tracking.}
\changed{{\em DukeMTMC-VideoReID} provides {\em manually} labelled trajectories,
originally introduced for {\em one-shot} person re-id \cite{wu2018exploit}. 
We tested UTAL on this dataset {\em without} the assumed 
one-shot labelled trajectory per ID.
We set $K\!=\!0$ for Eq \eqref{eq:sigma}
due to no trajectory fragmentation. 
Table \ref{tab:manual_track} shows that 
UTAL outperforms EUG \cite{wu2018exploit}
even without one-shot ID labelling.
This indicates the efficacy of our unsupervised learning strategy
in discovering re-id information.}

\begin{table}[h]
	\centering
	\setlength{\tabcolsep}{0.35cm}
	\caption{\changed{Evaluation on DukeMTMC-VideoReID.}}
	\label{tab:manual_track}
	\vskip -0.3cm
	\begin{tabular}
		{c||c|c|c|c}
		\hline
		Metric (\%) & Rank-1 & Rank-5 & Rank-20 & mAP
		\\ 
		\hline \hline
		EUG \cite{wu2018exploit}		& 72.8 & 84.2 & 91.5 & 63.2
		\\ \hline
		{\bf UTAL}		&\bf 74.5 &\bf  88.7 &\bf 96.3 &\bf 72.1
		\\ \hline
	\end{tabular}
	\vspace{-0.2cm}
\end{table}


\section{Conclusions}
We presented a novel {\em Unsupervised Tracklet Association Learning} (UTAL) model for
unsupervised tracklet person re-identification.
This model learns from
person tracklet data automatically extracted from videos, 
eliminating the
expensive and exhaustive manual ID labelling.
%
This enables UTAL to be more scalable to real-world applications. 
In contrast to existing re-id methods
that require exhaustively pairwise labelled training data for
every camera-pair or 
assume labelled source domain training data, 
the proposed UTAL model performs 
end-to-end deep learning of a person re-id model from scratch 
using totally unlabelled tracklet data. 
This is achieved by optimising jointly both
a Per-Camera Tracklet Discrimination loss function
and 
a Cross-Camera Tracklet Association loss function
in a unified architecture.
%
Extensive evaluations were conducted on eight 
image and video person re-id benchmarks 
to validate the advantages of
the proposed UTAL model over 
state-of-the-art 
unsupervised and domain adaptation re-id methods.


%

%

\ifCLASSOPTIONcompsoc
  \section*{Acknowledgments}
\else

\section*{Acknowledgment}
\fi
{\small This work is partially supported by 
Vision Semantics Limited,
National Natural Science Foundation of China (Project No. 61401212),
Royal Society Newton Advanced Fellowship Programme (NA150459),
and Innovate UK Industrial Challenge Project on Developing and Commercialising Intelligent Video Analytics Solutions for Public Safety (98111-571149).}

\ifCLASSOPTIONcaptionsoff
  \newpage
\fi



%

\bibliographystyle{IEEEtran}
\bibliography{IEEEabrv,mybibfile_abbr}

\begin{thebibliography}{10}
\providecommand{\url}[1]{#1}
\csname url@samestyle\endcsname
\providecommand{\newblock}{\relax}
\providecommand{\bibinfo}[2]{#2}
\providecommand{\BIBentrySTDinterwordspacing}{\spaceskip=0pt\relax}
\providecommand{\BIBentryALTinterwordstretchfactor}{4}
\providecommand{\BIBentryALTinterwordspacing}{\spaceskip=\fontdimen2\font plus
\BIBentryALTinterwordstretchfactor\fontdimen3\font minus
  \fontdimen4\font\relax}
\providecommand{\BIBforeignlanguage}[2]{{%
\expandafter\ifx\csname l@#1\endcsname\relax
\typeout{** WARNING: IEEEtran.bst: No hyphenation pattern has been}%
\typeout{** loaded for the language `#1'. Using the pattern for}%
\typeout{** the default language instead.}%
\else
\language=\csname l@#1\endcsname
\fi
#2}}
\providecommand{\BIBdecl}{\relax}
\BIBdecl

\bibitem{gong2014person}
S.~Gong, M.~Cristani, S.~Yan, and C.~C. Loy, \emph{Person
  re-identification}.\hskip 1em plus 0.5em minus 0.4em\relax Springer, 2014.

\bibitem{li2017person}
W.~Li, X.~Zhu, and S.~Gong, ``Person re-identification by deep joint learning
  of multi-loss classification,'' in \emph{Proc. Int. Jo. Conf. of Artif.
  Intell.}, 2017.

\bibitem{li2018harmonious}
------, ``Harmonious attention network for person re-identification,'' in
  \emph{Proc. IEEE Conf. Comput. Vis. Pattern Recognit.}, 2018, pp. 2285--2294.

\bibitem{wei2018person}
L.~Wei, S.~Zhang, W.~Gao, and Q.~Tian, ``Person transfer gan to bridge domain
  gap for person re-identification,'' in \emph{Proc. IEEE Conf. Comput. Vis.
  Pattern Recognit.}, 2018, pp. 79--88.

\bibitem{song2018mask}
C.~Song, Y.~Huang, W.~Ouyang, and L.~Wang, ``Mask-guided contrastive attention
  model for person re-identification,'' in \emph{Proc. IEEE Conf. Comput. Vis.
  Pattern Recognit.}, 2018, pp. 1179--1188.

\bibitem{chang2018multi}
X.~Chang, T.~M. Hospedales, and T.~Xiang, ``Multi-level factorisation net for
  person re-identification,'' in \emph{Proc. IEEE Conf. Comput. Vis. Pattern
  Recognit.}, 2018, pp. 2109--2118.

\bibitem{shen2018deep}
Y.~Shen, H.~Li, T.~Xiao, S.~Yi, D.~Chen, and X.~Wang, ``Deep group-shuffling
  random walk for person re-identification,'' in \emph{Proc. IEEE Conf. Comput.
  Vis. Pattern Recognit.}, 2018, pp. 2265--2274.

\bibitem{wang2014unsupervised}
H.~Wang, S.~Gong, and T.~Xiang, ``Unsupervised learning of generative topic
  saliency for person re-identification,'' in \emph{Proc. Bri. Mach. Vis.
  Conf.}, 2014.

\bibitem{kodirov2015dictionary}
E.~Kodirov, T.~Xiang, and S.~Gong, ``Dictionary learning with iterative
  laplacian regularisation for unsupervised person re-identification,'' in
  \emph{Proc. Bri. Mach. Vis. Conf.}, 2015.

\bibitem{lisanti2015person}
G.~Lisanti, I.~Masi, A.~D. Bagdanov, and A.~Del~Bimbo, ``Person
  re-identification by iterative re-weighted sparse ranking,'' \emph{IEEE
  Trans. Pattern Anal. Mach. Intell.}, vol.~37, no.~8, pp. 1629--1642, 2015.

\bibitem{kodirov2016person}
E.~Kodirov, T.~Xiang, Z.~Fu, and S.~Gong, ``Person re-identification by
  unsupervised $l_1$ graph learning,'' in \emph{Proc. Eur. Conf. Comput. Vis.},
  2016, pp. 178--195.

\bibitem{khan2016unsupervised}
F.~M. Khan and F.~Bremond, ``Unsupervised data association for metric learning
  in the context of multi-shot person re-identification,'' in \emph{Proc. IEEE
  Conf. Adv. Vid. Sig. Surv.}, 2016, pp. 256--262.

\bibitem{wang2016towards}
H.~Wang, X.~Zhu, T.~Xiang, and S.~Gong, ``Towards unsupervised open-set person
  re-identification,'' in \emph{IEEE Int. Conf. on Img. Proc.}, 2016, pp.
  769--773.

\bibitem{ma2017person}
X.~Ma, X.~Zhu, S.~Gong, X.~Xie, J.~Hu, K.-M. Lam, and Y.~Zhong, ``Person
  re-identification by unsupervised video matching,'' \emph{Pattern
  Recognition}, vol.~65, pp. 197--210, 2017.

\bibitem{ye2017dynamic}
M.~Ye, A.~J. Ma, L.~Zheng, J.~Li, and P.~C. Yuen, ``Dynamic label graph
  matching for unsupervised video re-identification,'' in \emph{Proc. IEEE Int.
  Conf. Comput. Vis.}, 2017, pp. 5142--5150.

\bibitem{liu2017stepwise}
Z.~Liu, D.~Wang, and H.~Lu, ``Stepwise metric promotion for unsupervised video
  person re-identification,'' in \emph{Proc. IEEE Int. Conf. Comput. Vis.},
  2017, pp. 2429--2438.

\bibitem{want2018Transfer}
J.~Wang, X.~Zhu, S.~Gong, and W.~Li, ``Transferable joint attribute-identity
  deep learning for unsupervised person re-identification,'' in \emph{Proc.
  IEEE Conf. Comput. Vis. Pattern Recognit.}, 2018, pp. 2275--2284.

\bibitem{peng2018joint}
P.~Peng, Y.~Tian, T.~Xiang, Y.~Wang, M.~Pontil, and T.~Huang, ``Joint semantic
  and latent attribute modelling for cross-class transfer learning,''
  \emph{IEEE Trans. Pattern Anal. Mach. Intell.}, vol.~40, no.~7, pp.
  1625--1638, 2018.

\bibitem{fan2017unsupervised}
H.~Fan, L.~Zheng, and Y.~Yang, ``Unsupervised person re-identification:
  Clustering and fine-tuning,'' \emph{arXiv:1705.10444}, 2017.

\bibitem{yu2017cross}
H.-X. Yu, A.~Wu, and W.-S. Zheng, ``Cross-view asymmetric metric learning for
  unsupervised person re-identification,'' in \emph{Proc. IEEE Int. Conf.
  Comput. Vis.}, 2017, pp. 994--1002.

\bibitem{li2014deepreid}
W.~Li, R.~Zhao, T.~Xiao, and X.~Wang, ``Deepreid: Deep filter pairing neural
  network for person re-identification,'' in \emph{Proc. IEEE Conf. Comput.
  Vis. Pattern Recognit.}, 2014, pp. 152--159.

\bibitem{zheng2015scalable}
L.~Zheng, L.~Shen, L.~Tian, S.~Wang, J.~Wang, and Q.~Tian, ``Scalable person
  re-identification: A benchmark,'' in \emph{Proc. IEEE Conf. Comput. Vis.
  Pattern Recognit.}, 2015, pp. 1116--1124.

\bibitem{ristani2016performance}
E.~Ristani, F.~Solera, R.~Zou, R.~Cucchiara, and C.~Tomasi, ``Performance
  measures and a data set for multi-target, multi-camera tracking,'' in
  \emph{Workshop of Eur. Conf. Comput. Vis.}, 2016, pp. 17--35.

\bibitem{zheng2017unlabeled}
Z.~Zheng, L.~Zheng, and Y.~Yang, ``Unlabeled samples generated by gan improve
  the person re-identification baseline in vitro,'' in \emph{Proc. IEEE Int.
  Conf. Comput. Vis.}, 2017, pp. 3754--3762.

\bibitem{wang2014person}
T.~Wang, S.~Gong, X.~Zhu, and S.~Wang, ``Person re-identification by video
  ranking,'' in \emph{Proc. Eur. Conf. Comput. Vis.}, 2014, pp. 688--703.

\bibitem{hirzer2011person}
M.~Hirzer, C.~Beleznai, P.~M. Roth, and H.~Bischof, ``Person re-identification
  by descriptive and discriminative classification,'' in \emph{Scand. Conf.
  Img. Anal.}, 2011, pp. 91--102.

\bibitem{zheng2016mars}
L.~Zheng, Z.~Bie, Y.~Sun, J.~Wang, C.~Su, S.~Wang, and Q.~Tian, ``Mars: A video
  benchmark for large-scale person re-identification,'' in \emph{Proc. Eur.
  Conf. Comput. Vis.}, 2016, pp. 868--884.

\bibitem{li2018Unsupervised}
M.~Li, X.~Zhu, and S.~Gong, ``Unsupervised person re-identification by deep
  learning tracklet association,'' in \emph{Proc. Eur. Conf. Comput. Vis.},
  2018, pp. 737--753.

\bibitem{chen2018person}
Y.-C. Chen, X.~Zhu, W.-S. Zheng, and J.-H. Lai, ``Person re-identification by
  camera correlation aware feature augmentation,'' \emph{IEEE Trans. Pattern
  Anal. Mach. Intell.}, vol.~40, no.~2, pp. 392--408, 2018.

\bibitem{wang2016human}
H.~Wang, S.~Gong, X.~Zhu, and T.~Xiang, ``Human-in-the-loop person
  re-identification,'' in \emph{Proc. Eur. Conf. Comput. Vis.}, 2016, pp.
  405--422.

\bibitem{liu2013pop}
C.~Liu, C.~Change~Loy, S.~Gong, and G.~Wang, ``Pop: Person re-identification
  post-rank optimisation,'' in \emph{Proc. IEEE Int. Conf. Comput. Vis.}, 2013,
  pp. 441--448.

\bibitem{farenzena2010person}
M.~Farenzena, L.~Bazzani, A.~Perina, V.~Murino, and M.~Cristani, ``Person
  re-identification by symmetry-driven accumulation of local features,'' in
  \emph{Proc. IEEE Conf. Comput. Vis. Pattern Recognit.}, 2010, pp. 2360--2367.

\bibitem{zhao2013unsupervised}
R.~Zhao, W.~Ouyang, and X.~Wang, ``Unsupervised salience learning for person
  re-identification,'' in \emph{Proc. IEEE Conf. Comput. Vis. Pattern
  Recognit.}, 2013, pp. 3586--3593.

\bibitem{liu2014semi}
X.~Liu, M.~Song, D.~Tao, X.~Zhou, C.~Chen, and J.~Bu, ``Semi-supervised coupled
  dictionary learning for person re-identification,'' in \emph{Proc. IEEE Conf.
  Comput. Vis. Pattern Recognit.}, 2014, pp. 3550--3557.

\bibitem{zhu2017unpaired}
J.-Y. Zhu, T.~Park, P.~Isola, and A.~A. Efros, ``Unpaired image-to-image
  translation using cycle-consistent adversarial networks,'' in \emph{Proc.
  IEEE Int. Conf. Comput. Vis.}, 2017, pp. 2223--2232.

\bibitem{deng2018image}
W.~Deng, L.~Zheng, Q.~Ye, G.~Kang, Y.~Yang, and J.~Jiao, ``Image-image domain
  adaptation with preserved self-similarity and domain-dissimilarity for person
  re-identification,'' in \emph{Proc. IEEE Conf. Comput. Vis. Pattern
  Recognit.}, 2018, pp. 994--1003.

\bibitem{zhong2018generalizing}
Z.~Zhong, L.~Zheng, S.~Li, and Y.~Yang, ``Generalizing a person retrieval model
  hetero-and homogeneously,'' in \emph{Proc. Eur. Conf. Comput. Vis.}, 2018,
  pp. 172--188.

\bibitem{peng2016unsupervised}
P.~Peng, T.~Xiang, Y.~Wang, M.~Pontil, S.~Gong, T.~Huang, and Y.~Tian,
  ``Unsupervised cross-dataset transfer learning for person
  re-identification,'' in \emph{Proc. IEEE Conf. Comput. Vis. Pattern
  Recognit.}, 2016, pp. 1306--1315.

\bibitem{lv2018unsupervised}
J.~Lv, W.~Chen, Q.~Li, and C.~Yang, ``Unsupervised cross-dataset person
  re-identification by transfer learning of spatial-temporal patterns,'' in
  \emph{Proc. IEEE Conf. Comput. Vis. Pattern Recognit.}, 2018, pp. 7948--7956.

\bibitem{bak2018domain}
S.~Bak, P.~Carr, and J.-F. Lalonde, ``Domain adaptation through synthesis for
  unsupervised person re-identification,'' in \emph{Proc. Eur. Conf. Comput.
  Vis.}, 2018, pp. 189--205.

\bibitem{argyriou2007multi}
A.~Argyriou, T.~Evgeniou, and M.~Pontil, ``Multi-task feature learning,'' in
  \emph{Proc. Neur. Info. Proc. Sys.}, 2007, pp. 41--48.

\bibitem{caruana1997multitask}
R.~Caruana, ``Multitask learning,'' \emph{Mach. Learn.}, vol.~28, no.~1, pp.
  41--75, 1997.

\bibitem{dong2017multi}
Q.~Dong, S.~Gong, and X.~Zhu, ``Multi-task curriculum transfer deep learning of
  clothing attributes,'' in \emph{Proc. IEEE Win. Conf. App. of Comp. Vis.},
  2017, pp. 520--529.

\bibitem{zhang2016learning}
Z.~Zhang, P.~Luo, C.~C. Loy, and X.~Tang, ``Learning deep representation for
  face alignment with auxiliary attributes,'' \emph{IEEE Trans. Pattern Anal.
  Mach. Intell.}, vol.~38, no.~5, pp. 918--930, 2016.

\bibitem{olshausen1997sparse}
B.~A. Olshausen and D.~J. Field, ``Sparse coding with an overcomplete basis
  set: A strategy employed by v1?'' \emph{Vision Research}, vol.~37, no.~23,
  pp. 3311--3325, 1997.

\bibitem{xie2016unsupervised}
J.~Xie, R.~Girshick, and A.~Farhadi, ``Unsupervised deep embedding for
  clustering analysis,'' in \emph{Proc. Int. Conf. Mach. Learn.}, 2016, pp.
  478--487.

\bibitem{bojanowski2017unsupervised}
P.~Bojanowski and A.~Joulin, ``Unsupervised learning by predicting noise,'' in
  \emph{Proc. Int. Conf. Mach. Learn.}, 2017, pp. 517--526.

\bibitem{wu2018unsupervised}
Z.~Wu, Y.~Xiong, X.~Y. Stella, and D.~Lin, ``Unsupervised feature learning via
  non-parametric instance discrimination,'' in \emph{Proc. IEEE Conf. Comput.
  Vis. Pattern Recognit.}, 2018, pp. 3733--3742.

\bibitem{wang2015unsupervised}
X.~Wang and A.~Gupta, ``Unsupervised learning of visual representations using
  videos,'' in \emph{Proc. IEEE Int. Conf. Comput. Vis.}, 2015, pp. 2794--2802.

\bibitem{liu2016ssd}
W.~Liu, D.~Anguelov, D.~Erhan, C.~Szegedy, S.~Reed, C.-Y. Fu, and A.~C. Berg,
  ``Ssd: Single shot multibox detector,'' in \emph{Proc. Eur. Conf. Comput.
  Vis.}, 2016, pp. 21--37.

\bibitem{zhang2016far}
S.~Zhang, R.~Benenson, M.~Omran, J.~Hosang, and B.~Schiele, ``How far are we
  from solving pedestrian detection?'' in \emph{Proc. IEEE Conf. Comput. Vis.
  Pattern Recognit.}, 2016, pp. 1259--1267.

\bibitem{leal2015motchallenge}
L.~Leal-Taix{\'e}, A.~Milan, I.~Reid, S.~Roth, and K.~Schindler, ``Motchallenge
  2015: Towards a benchmark for multi-target tracking,''
  \emph{arXiv:1504.01942}, 2015.

\bibitem{belkin2006manifold}
M.~Belkin, P.~Niyogi, and V.~Sindhwani, ``Manifold regularization: A geometric
  framework for learning from labeled and unlabeled examples,'' \emph{Journ. of
  Mach. Learn. Res.}, vol.~7, pp. 2399--2434, 2006.

\bibitem{zelnik2005self}
L.~Zelnik-Manor and P.~Perona, ``Self-tuning spectral clustering,'' in
  \emph{Proc. Neur. Info. Proc. Sys.}, 2005, pp. 1601--1608.

\bibitem{hinton2015distilling}
G.~Hinton, O.~Vinyals, and J.~Dean, ``Distilling the knowledge in a neural
  network,'' \emph{arXiv preprint arXiv:1503.02531}, 2015.

\bibitem{zhang2016understanding}
C.~Zhang, S.~Bengio, M.~Hardt, B.~Recht, and O.~Vinyals, ``Understanding deep
  learning requires rethinking generalization,'' in \emph{Proc. Int. Conf. on
  Learn. Rep.}, 2017.

\bibitem{qin2011hello}
D.~Qin, S.~Gammeter, L.~Bossard, T.~Quack, and L.~Van~Gool, ``Hello neighbor:
  Accurate object retrieval with k-reciprocal nearest neighbors,'' in
  \emph{Proc. IEEE Conf. Comput. Vis. Pattern Recognit.}, 2011, pp. 777--784.

\bibitem{bengio2009curriculum}
Y.~Bengio, J.~Louradour, R.~Collobert, and J.~Weston, ``Curriculum learning,''
  in \emph{Proc. Int. Conf. Mach. Learn.}, 2009, pp. 41--48.

\bibitem{hardoon2004canonical}
D.~R. Hardoon, S.~Szedmak, and J.~Shawe-Taylor, ``Canonical correlation
  analysis: An overview with application to learning methods,'' \emph{Neural
  Computation}, vol.~16, no.~12, pp. 2639--2664, 2004.

\bibitem{schroff2015facenet}
F.~Schroff, D.~Kalenichenko, and J.~Philbin, ``Facenet: A unified embedding for
  face recognition and clustering,'' in \emph{Proc. IEEE Conf. Comput. Vis.
  Pattern Recognit.}, 2015, pp. 815--823.

\bibitem{lee2008sparse}
H.~Lee, C.~Ekanadham, and A.~Y. Ng, ``Sparse deep belief net model for visual
  area v2,'' in \emph{Proc. Neur. Info. Proc. Sys.}, 2008, pp. 873--880.

\bibitem{xiao2016learning}
T.~Xiao, H.~Li, W.~Ouyang, and X.~Wang, ``Learning deep feature representations
  with domain guided dropout for person re-identification,'' in \emph{Proc.
  IEEE Conf. Comput. Vis. Pattern Recognit.}, 2016, pp. 1249--1258.

\bibitem{ye2007adaptive}
J.~Ye, Z.~Zhao, and H.~Liu, ``Adaptive distance metric learning for
  clustering,'' in \emph{Proc. IEEE Conf. Comput. Vis. Pattern Recognit.},
  2007, pp. 1--7.

\bibitem{qin2015unsupervised}
C.~Qin, S.~Song, G.~Huang, and L.~Zhu, ``Unsupervised neighborhood component
  analysis for clustering,'' \emph{Neurocomputing}, vol. 168, pp. 609--617,
  2015.

\bibitem{chen2018group}
D.~Chen, D.~Xu, H.~Li, N.~Sebe, and X.~Wang, ``Group consistent similarity
  learning via deep crf for person re-identification,'' in \emph{Proc. IEEE
  Conf. Comput. Vis. Pattern Recognit.}, 2018, pp. 8649--8658.

\bibitem{wu2018exploit}
Y.~Wu, Y.~Lin, X.~Dong, Y.~Yan, W.~Ouyang, and Y.~Yang, ``Exploit the unknown
  gradually: One-shot video-based person re-identification by stepwise
  learning,'' in \emph{Proc. IEEE Conf. Comput. Vis. Pattern Recognit.}, 2018,
  pp. 5177--5186.

\bibitem{gou2017dukemtmc4reid}
M.~Gou, S.~Karanam, W.~Liu, O.~Camps, and R.~J. Radke, ``Dukemtmc4reid: A
  large-scale multi-camera person re-identification dataset,'' in
  \emph{Workshop of IEEE Conf. Comput. Vis. Pattern Recognit.}, 2017, pp.
  10--19.

\bibitem{szegedy2015going}
C.~Szegedy, W.~Liu, Y.~Jia, P.~Sermanet, S.~Reed, D.~Anguelov, D.~Erhan,
  V.~Vanhoucke, and A.~Rabinovich, ``Going deeper with convolutions,'' in
  \emph{Proc. IEEE Conf. Comput. Vis. Pattern Recognit.}, 2015, pp. 1--9.

\bibitem{he2016deep}
K.~He, X.~Zhang, S.~Ren, and J.~Sun, ``Deep residual learning for image
  recognition,'' in \emph{Proc. IEEE Conf. Comput. Vis. Pattern Recognit.},
  2016, pp. 770--778.

\bibitem{kingma2014adam}
D.~P. Kingma and J.~Ba, ``Adam: A method for stochastic optimization,''
  \emph{arXiv:1412.6980}, 2014.

\bibitem{ye2018robust}
M.~Ye, X.~Lan, and P.~C. Yuen, ``Robust anchor embedding for unsupervised video
  person re-identification in the wild,'' in \emph{Proc. Eur. Conf. Comput.
  Vis.}, 2018, pp. 170--186.

\bibitem{chen2018deep}
Y.~Chen, X.~Zhu, and S.~Gong, ``Deep association learning for unsupervised
  video person re-identification,'' \emph{Proc. Bri. Mach. Vis. Conf.}, 2018.

\bibitem{chen2018video}
D.~Chen, H.~Li, T.~Xiao, S.~Yi, and X.~Wang, ``Video person re-identification
  with competitive snippet-similarity aggregation and co-attentive snippet
  embedding,'' in \emph{Proc. IEEE Conf. Comput. Vis. Pattern Recognit.}, 2018,
  pp. 1169--1178.

\bibitem{li2012human}
W.~Li, R.~Zhao, and X.~Wang, ``Human reidentification with transferred metric
  learning,'' in \emph{Proc. Asi. Conf. Comp. Vis.}, 2012, pp. 31--44.

\bibitem{gray2008viewpoint}
D.~Gray and H.~Tao, ``Viewpoint invariant pedestrian recognition with an
  ensemble of localized features,'' in \emph{Proc. Eur. Conf. Comput. Vis.},
  2008, pp. 262--275.

\bibitem{prosser2010person}
B.~J. Prosser, W.-S. Zheng, S.~Gong, and T.~Xiang, ``Person re-identification
  by support vector ranking,'' in \emph{Proc. Bri. Mach. Vis. Conf.}, 2010.

\bibitem{pearce2005improved}
D.~J. Pearce, ``An improved algorithm for finding the strongly connected
  components of a directed graph,'' \emph{Victoria University, Wellington, NZ,
  Tech. Rep}, 2005.

\bibitem{huang2017densely}
G.~Huang, Z.~Liu, L.~Van Der~Maaten, and K.~Q. Weinberger, ``Densely connected
  convolutional networks,'' in \emph{Proc. IEEE Conf. Comput. Vis. Pattern
  Recognit.}, 2017, pp. 4700--4708.

\end{thebibliography}


%

\vspace{-1.5cm}
\begin{IEEEbiography}[{\includegraphics[width=1in,height=1.25in,clip,keepaspectratio]{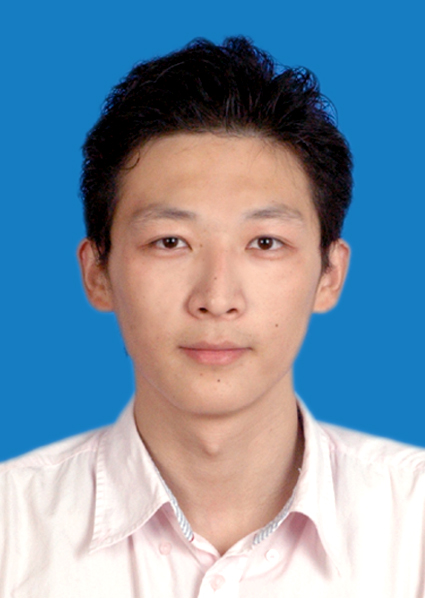}}]{Minxian Li}
is a postdoctoral researcher at Queen Mary University of London, United Kindom and also an assistant professor of Nanjing University of Science and Technology, China.
He received his Ph.D. in Pattern Recognition and Intelligent System
from Nanjing University of Science and Technology, China.
His research interests include computer vision, pattern recognition and deep learning.
\end{IEEEbiography}

\vspace{-1.5cm}
\begin{IEEEbiography}[{\includegraphics[width=1in,height=1.25in,clip,keepaspectratio]{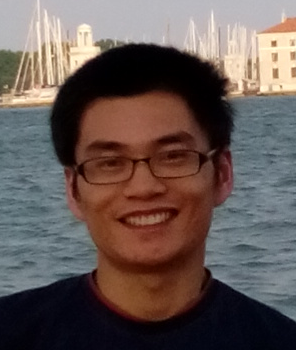}}]{Xiatian Zhu}
is a Computer Vision Researcher at Vision Semantics Limited, London, UK.
He received his Ph.D. from Queen Mary University of London. 
He won The Sullivan Doctoral Thesis Prize 2016,
an annual award representing the best doctoral thesis 
submitted to a UK University in computer vision.
His research interests include computer vision and machine learning.
\end{IEEEbiography}

\vspace{-1.5cm}
\begin{IEEEbiography}[{\includegraphics[width=1in,height=1.25in,clip,keepaspectratio]{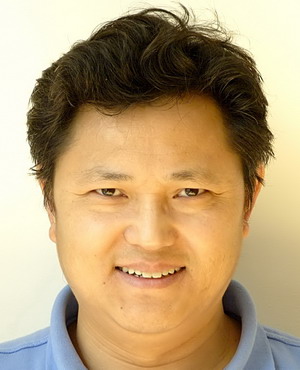}}]{Shaogang Gong}
is Professor of Visual Computation at Queen Mary University of London (since 2001), 
a Fellow of the Institution of Electrical Engineers and a Fellow of the British Computer Society. 
He received his D.Phil (1989) in computer vision from Keble College, Oxford University. 
His research interests include computer vision, machine learning and video analysis.
\end{IEEEbiography}







\end{document}